%% file: revised_version_1.tex
\documentclass[10pt,journal,compsoc]{IEEEtran}
\usepackage{amsmath,amsfonts}
\usepackage{array}
\usepackage{textcomp}
\usepackage{stfloats}
\usepackage{url}
\usepackage{verbatim}
\usepackage{graphicx}
\usepackage{cite}
\usepackage[justification=centering]{caption}
\usepackage{subcaption}

\usepackage{wrapfig}
\usepackage[outdir=./]{epstopdf}
\usepackage{multirow}
\usepackage{makecell}
\usepackage{xcolor}
\usepackage{makecell}
\usepackage{enumitem}

\usepackage{graphicx}
\usepackage{gensymb}
\usepackage{amsmath}

\usepackage{amssymb}
\usepackage{stmaryrd}
\usepackage{amsthm} 
\usepackage[ruled,vlined]{algorithm2e}
\usepackage{algpseudocode}
\usepackage{varwidth}
\usepackage{blindtext}
\usepackage{scrextend}
\usepackage{url}
\usepackage[outdir=./]{epstopdf}

\usepackage[ruled,vlined]{algorithm2e}

\usepackage[square,comma,numbers,sort]{natbib}

\usepackage{mathtools}

\usepackage[switch]{lineno}

\DeclarePairedDelimiter\floor{\lfloor}{\rfloor}

\theoremstyle{plain}

\theoremstyle{definition}

\makeatletter
\newcommand{\circlearrow}{}
\DeclareRobustCommand{\circlearrow}{%
  \mathrel{\vphantom{\rightarrow}\mathpalette\circle@arrow\relax}%
}
\newcommand{\circle@arrow}[2]{%
  \m@th
  \ooalign{%
    \hidewidth$#1\circ\mkern1mu$\hidewidth\cr
    $#1\longrightarrow$\cr}%
}
\makeatother

\usepackage{enumitem}

\begin{document}

\title{Real-Time Sense and Detect of Drones Using Deep Learning and Airborne LiDAR}

\author{Manduhu Manduhu, Alexander Dow, Petar Trslic, Gerard Dooly, Benjamin Blanck, James Riordan 
       
\thanks{
	Manduhu Manduhu, James Riordan, and Alexander Dow are with the Drone Systems Lab, School of Computing, Engineering, and Physical Sciences, University of the West of Scotland, Glasgow, Scotland. 

	Corresponding author: James Riordan (email: james.riordan@uws.ac.uk). 
	
	Petar Trslic and Gerard Dooly are with the Centre for Robotics \& Intelligent Systems, Department of Electronic \& Computer Engineering, University of Limerick, Limerick, Ireland.
	
	Benjamin Blanck is with Hamburg Port Authority, Hamburg, Germany.
	
	}
}

\markboth{Journal of \LaTeX\ Class Files,~Vol.~xx, No.~xx, xxxx~20xx}%
{Shell \MakeLowercase{\textit{et al.}}: A Sample Article Using IEEEtran.cls for IEEE Journals}


\maketitle

\input {sec/abstract.tex}

\input {sec/motivation.tex}

\input {sec/related-work.tex}

\input {sec/method.tex}

\input {sec/training.tex}
\input {sec/testing.tex}
\input {sec/experimental-system.tex}

\input {sec/experimental-result.tex}

\input {sec/discusstion.tex}

\input {sec/conclusion.tex}
\input {sec/bio.tex}

\newpage
\appendices
\input {sec/appendix.tex}

\end{document}

%% file: sec/abstract.tex
\begin{abstract}
The safe operation of drone swarms beyond visual line of sight requires multiple safeguards to mitigate the risk of collision between drones flying in close-proximity   scenarios. Cooperative navigation and flight coordination strategies that rely on pre-planned trajectories, constant 
network connectivity and reliable Global Navigation Satellite System (GNSS) positioning are brittle to failure. Drone embedded sense and detect
offers a comprehensive mode of
separation between drones for deconfliction and collision avoidance. 
This paper presents the first airborne LiDAR-based solution for drone-swarm detection and localization using a 3D deep learning model. 
It adapts an existing deep learning neural network to the air-to-air drone scenario by expanding the scan space vertically.
A new sparse convolution is proposed and applied to accelerate the backbone layer, which is the most time-consuming part of the neural network.
To collect training data of safety critical, close-proximity multi-drone operations, a scenario Digital Twin is used to augment real datasets with high-fidelity synthetic data. The trained model achieves over 80{\%} recall and 96{\%} precision when tested on real-world datasets. 
\end{abstract}

\begin{IEEEkeywords}
sense-and-detect, drone swarm, deep learning, sparse convolution, digital twin. 
\end{IEEEkeywords}

%% file: sec/motivation.tex
\section{Introduction}
\IEEEPARstart{S}{ense} and detect algorithms which involve identifying potential conflicts, are crucial for assessing risks in dynamic drone swarm environments \cite{IEEEsenseavoid}.
In the context of drone conflict avoidance, sense and detect can be defined as follows:
Sense refers to the methods employed by a drone to actively gather information about its surrounding environment. These methods typically involve onboard sensor technologies, such as cameras, LiDAR, and other advanced sensing systems.
Detect involves analyzing the data collected by the drone's sensors. The analysis aims to identify the presence of threats and hazards within the drone's operational space and assess the potential and severity of these threats to the drone's safe operation.
Together, these functions allow the drone to recognize obstacles and assess risks in real time, forming the foundation for effective avoidance strategies.


As drone use cases expand, it becomes essential to validate technologies that can take on the safety-critical role of onboard pilots by monitoring surrounding airspace for hazards, including air and ground collision risks. Current drone swarm management systems rely primarily on trajectory-based flight planning, where predefined flight paths guide drones to avoid conflicts. Simultaneous Localization and Mapping (SLAM), which enables drones to build a map of their environment and track their position within it, along with anti-spoofing techniques that protect against falsified GPS signals, have improved in-flight positional estimation and robustness. However, tactical deconfliction—the process of managing and resolving potential flight path conflicts—still depends on centralized decision-making and continuous connectivity to a command and control (C2) center. This centralized system is vulnerable to network failures, and it relies on cooperative behavior from all drones. To safeguard users and mitigate air-to-air collision risks, onboard drone-embedded sense and detect capabilities, where drones autonomously perceive and avoid conflicts, are necessary.

While ensuring safe separation in drone swarms is critical for collision avoidance, these capabilities also enable novel applications beyond airspace management. One such application is the inspection of critical infrastructure, where autonomous drone swarms can significantly enhance efficiency and coverage  \cite{Gkoumas02082024}\cite{10399955}\cite{thales}. 
Multiple civil engineering reports state that 1 in 10 bridges are at risk of failure \cite{adam2024risks}\cite{ali2019artificial}, underscoring the need for advanced inspection solutions. Our research demonstrates that drone swarms can enable continuous large-scale infrastructure inspection and monitoring in complex transport system environments, reducing operational downtime and enhancing coverage efficiency. In our drone swarm framework, a monitor drone utilizes LiDAR-based 3D object detection to track and coordinate the swarm, mitigating collision risks caused by GNSS drift \cite{GPSdrift} and localization errors. The approach improves the safety, reliability, and efficiency of inspections. This paper focuses on the 3D point cloud detection model. Our contributions are:

\begin{enumerate}
\item We present a real-time airborne LiDAR-based solution for drone-swarm detection and localization using 3D deep learning, designed to operate in complex transportation infrastructure environments. The model introduces altitude-stratified anchor boxes to improve detection robustness for airborne objects at varying flight levels, enabling effective sense-and-detect capabilities in high-risk swarm scenarios.
\item A novel sparse convolution algorithm is developed to achieve real-time inference on airborne embedded platforms, leveraging a scattering-based implementation that eliminates the need for hash tables or rule books. This optimization significantly reduces inference latency, enabling deployment on edge hardware while maintaining detection accuracy. The algorithm accelerates the detection model, achieving a 2.3× speedup.
\item  A scenario-driven Digital Twin is used to generate high-fidelity synthetic training data, simulating LiDAR scans of drones in diverse safety-critical scenarios. This blended reality augmentation approach enhances model performance, achieving 80\% recall and 96\% precision, compared to 62\% recall and 82.6\% precision when trained solely on real-world data.
\end{enumerate}

%% file: sec/related-work.tex
\section{Related Work} \label{ref}
In drone conflict avoidance, a drone gathers environmental data using onboard sensors such as cameras and LiDAR, then analyzes this data to identify potential threats and assess their severity to ensure safe operation.

\subsection{Detection and Tracking of Drones}
\subsubsection{Multisensor Approaches for Drone Detection}

A survey \cite{hommes2016detection} is conducted to examine the distinctive marks of small drones utilizing a variety of sensor technologies. These technologies include acoustical antennas, passive and active optical imaging tools, as well as compact FMCW RADAR systems.  As described in \cite{acoustic1}, acoustic-based detection of drones utilizing various machine learning algorithms offers a promising approach for drone identification. However, this method inherently faces limitations in precise localization in 3D space.
While \cite{RF1} presents a method for single-drone detection using passive radio frequency (RF) signals, this technique is less effective for drone swarms. The presence of multiple drones operating in the same area 
elevates the overall background RF noise level, hindering the ability of RF-based detection systems to isolate and identify the weaker signals from individual drones within the swarm. 

RADAR technology \cite{radar1}\cite{radar2}\cite{radarHector} presents a valuable approach for drone detection, though its effectiveness is hampered by the low radar cross-section (RCS) of drones. This limitation can lead to missed detections, particularly for smaller drones, significantly restricting its use in safety-critical applications
of drone swarm monitoring. 
Several camera-based detect and track of small drones are presented in \cite{aker2017using}\cite{Detec-small-UVAs-pami}\cite{unlu2019deep}\cite{detectFromVideo}.
Camera-based drone detection offers an efficient solution for detection and classification of drones, however, their performance is significantly impacted by environmental factors like lighting conditions and weather.
Additionally, cameras inherently provide only 2D information, lacking the crucial depth data required for precise 3D localization. While stereo vision \cite{StereoVisionDepth} can offer depth estimates, accurate depth for larger distances requires prohibitively long stereo baselines, which can be challenging for our needs.


A complete chain of detection, tracking, and classification of small flying objects in real-time is presented in \cite{Hammer2020AMA}. This is achieved through the use of a mobile multi-sensor platform equipped with two 360$^{\circ}$ LiDAR scanners and pan-and-tilt cameras in the visible and thermal infrared (IR) spectrum. The approach involves the initial detect and track of flying objects in 3D LiDAR data. After localizing, the cameras are automatically directed to the object's position, and each sensor records a 2D image. A convolutional neural network (CNN) is then used to identify the region of interest (ROI) and classify the object into one of eight different types of drones and birds. Clearly this two-stage detection and classification approach will introduce higher latency. Alongside latency issues, deploying this system on a drone poses further challenges due to the weight and power limitations inherent in drone platforms.

As drones have gained popularity, the attention towards cooperative drone detection and tracking has intensified.
To improve cooperative drone navigation in challenging GNSS conditions, the approach presented in \cite{causa2022closed} combines camera based measurements with navigation data from other aircraft flying under better GNSS coverage. Clearly, this approach is not independent of external positioning systems such as GNSS which can introduce multipath errors caused by signals reflecting off buildings, trees, the ground, and other surfaces.

This paper presents an airborne LiDAR based solution for drone detection using a point cloud-based 3D deep learning network. 
This approach represents an advancement in drone detection technology, 
as it satisfies different requirements for real-time sense and avoid of drones in safety critical scenarios. 
These requirements include low latency, high detection performance, 3D localization capability, independence from external positioning systems, and the ability to operate within the strict power and weight limitations of drones.

\subsubsection{Deep learning on points clouds}

Recently deep learning on point clouds has become a prevalent AI technique in autonomous driving. Numerous methods have been proposed to address different problems in this area. A comprehensive review of deep learning approaches on point clouds is presented in \cite{PC-Survey-2021}. 
As a pioneering work on point clouds classification and segmentation, PointNet \cite{PointNet} and its successor PointNet++ \cite{PointNet++} have inspired many other works in this area.

VoxelNet \cite{Voxelnet} organizes point clouds into evenly spaced 3D voxels and applies a PointNet to encode point-wise features within each voxel. A 3D convolutional middle layer and a region proposal network are then applied to generate 3D object proposals in the form of 3D bounding boxes. 
The computational bottleneck resides in the 3D convolution layer. SECOND \cite{second} improved VoxelNet by applying sparse convolution and achieved an inference speed of 30.4 FPS \cite{deng2021voxel} on a NVIDIA RTX 2080 Ti GPU,  but the 3D convolutions remain a bottleneck.
PointPillars \cite{PointPillars} enables end-to-end learning with only 2D convolutional layers. It divides the x-y coordinate plane into evenly spaced grid cells and organizes point clouds into the corresponding cell to create ”pillars”. A simplified PointNet is applied to the points within pillars. A  backbone network consisting of multiple 2D convolutions is then applied to pseudo-image of features generated from the processed pillars. A detection head \cite{SSD} is then used to generate 3D bounding box for objects in point cloud data. 
The inference speed of PointPillars reaches 42.4 FPS, making it highly suitable for applications that demand real-time processing.


PV-RCNN \cite{shi2020pvrcnn} extends SECOND by adding a keypoints branch to preserve 3D structural information.
Its backbone network takes the raw point cloud data as input to extract point-wise features. PV-RCNN can achieve a higher accuracy but its inference speed is only 8.9 FPS, which falls below real-time requirements.
Voxel-RCNN \cite{deng2021voxel} further improves accuracy by leveraging a 3D network for feature extraction. It then applies a 2D backbone network and Region Proposal Network to BEV representation of the 3D feature volumes. Voxel-RCNN achieves an inference speed of 25.2 FPS.


CenterPoint \cite{yin2021center} reformulates detection as a center-based approach rather than relying on anchors or dense bounding box predictions. This design improves both detection quality and inference speed, achieving 25.0 FPS on the KITTI dataset.
An attention-based model, DSVT, is proposed in \cite{DSVT}, which selectively processes only the important sparse voxel regions, improving both efficiency and scalability. With this approach, DSVT achieves both high accuracy and fast inference, delivering a runtime of 15.9 FPS on the KITTI dataset.
CenterPoint and DSVT provide both voxel-based and pillar-based models. The inference speed of different models is summarized in Table \ref{tab:diff-model-perf}.

\begin{table}[ht]
\centering
\caption{Inference Speed of Different Models on KITTI Dataset using NVIDIA RTX 2080 Ti GPU}
\begin{tabular}{c|c}
\hline
\textbf{Model} & \textbf{FPS}  \\ \hline \hline
SECOND & 30.4 \\ \hline
PointPillars & 42.4  \\ \hline
PV-RCNN & 8.9  \\ \hline
Voxel-RCNN &25.2 \\ \hline
CenterPoint & 25.0  \\ \hline
DSVT  & 15.9 \\ \hline
\end{tabular}
\label{tab:diff-model-perf}
\end{table}

While there are various 3D object detection networks that can achieve higher detection performance, we selected PointPillars for our work due to its real-time inference capability; it consistently achieves the highest FPS compared to other detection models \cite{deng2021voxel}.
This is a critical consideration for our safety-oriented approach, which requires the ability to process incoming point clouds in real time.
While higher detection performance is desirable, it often increases computation time and resource demands, potentially hindering real-world deployment. We chose PointPillars as it can fulfill the real-time requirements of our drone detection applications.


Instead of purely using the standard PointPillars network, we modify it for an air-to-air drone scenario by extending the scan space vertically. 
To reduce inference latency, we propose a new sparse convolution algorithm to accelerate the backbone layer, which is the most time-consuming part of PointPillars \cite{backbonetypeselection}.

\begin{figure*}[ht]
\centering
\includegraphics[scale=0.43]{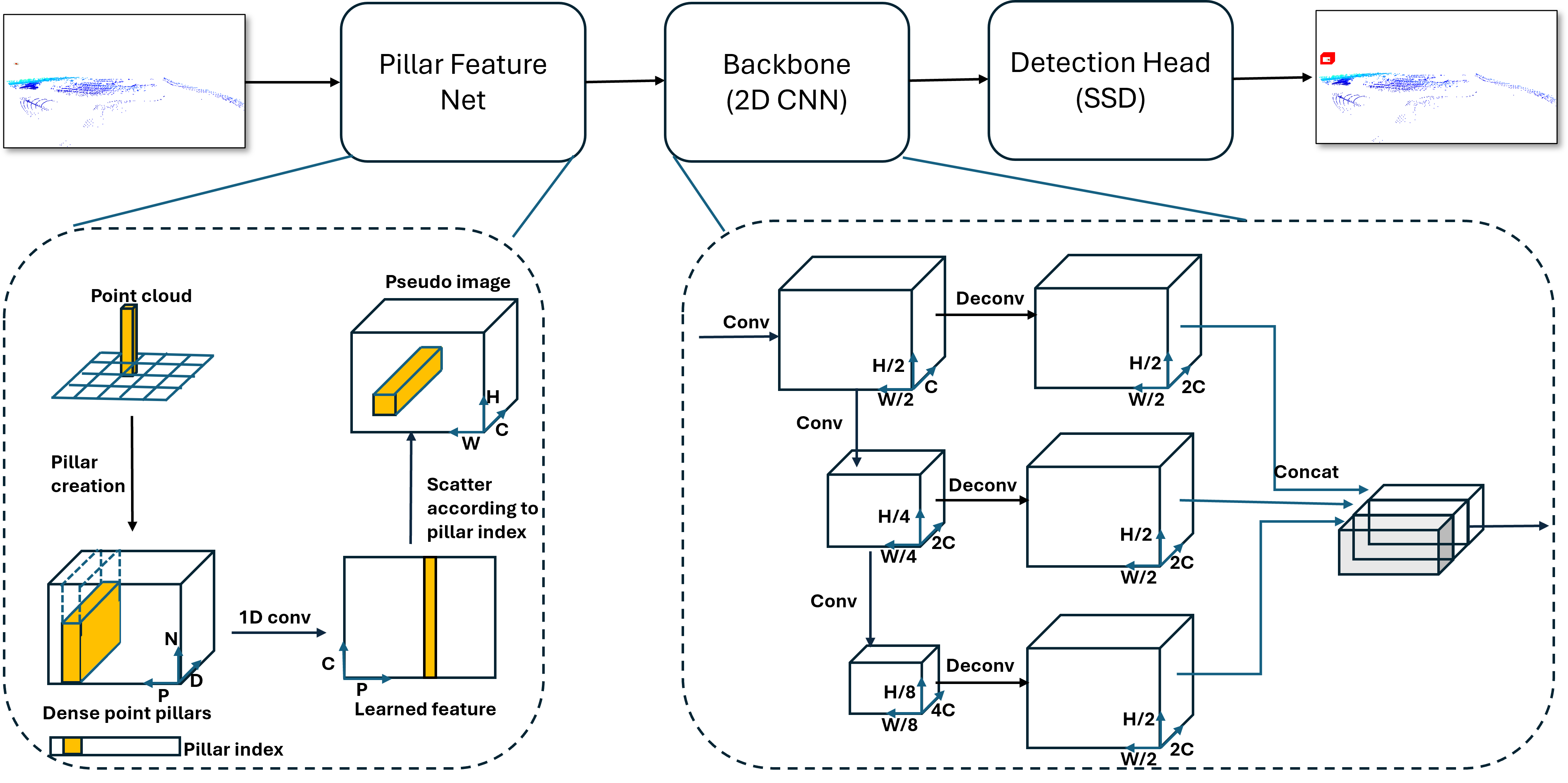}
\caption{The PointPillars architecture consists of these three key building blocks: Pillar Feature Net, Backbone, and Detection Head.}
\label{fig:pointpillar-arch}
\end{figure*}

\subsection{Latency Reduction}\label{sec:sparseworks}
While increasing point cloud density can improve detection accuracy, minimizing latency is crucial for ensuring real-time hazard response.
The inherent sparsity of LiDAR data can be utilized to reduce latency.
The study in \cite{graham2014spatially} describes the spatial sparsity of feature maps and how sparsity applies to improve performance of different applications.
A sparse convolutional neural network that utilizes sparse matrix multiplication is presented in \cite{spcnn}.
An efficient sparse convolutional operation called submanifold sparse convolution is introduced in \cite{Submanifold}, in which the output will be computed if and only if the corresponding site in the input is active. 

Matrix multiplication is an efficient method for implementing spatial convolution on GPU, as demonstrated by \cite{cudnn-arxiv}. 
This requires transforming the input image into matrices that are suitable for fast multiplication.
This transformation can be achieved by lowering the input data with duplication.
The GPU implementation of the proposed method in this paper is also based on fast GPU matrix multiplication, but without data lowering. 


A sparse convolution with rule book is presented in \cite{second}, where the rule book is used to map between input and output.
The work presented in \cite{minkowski} generalizes convolution, including sparse convolution, to handle any discrete convolutions in high-dimensional spaces.
A highly efficient Sparse Kernel Generator is developed in \cite{torchsparse}, which produces high performance sparse point cloud convolution kernels at a fraction of the engineering cost of the current state-of-the-art. 
In \cite{deltacnn}, the computational efficiency is improved by applying convolutions solely to the changed regions within successive video frames, enabling sparse feature updates to propagate throughout the network.

Most current methods implement convolution as a gathering operation, where neighboring pixels are multiplied by corresponding filter elements, and the results are summed to update the center pixel. Sparse convolution implementations typically follow this approach as well. To accelerate GPU performance, a hash table or rule book is often used to map positions between the input and output.
In this paper, we propose an alternative approach to implement sparse convolution using a scattering operation. This method eliminates the need for a hash table or rule book, improving efficiency while simplifying the implementation.

\subsection{Training Data Generation}
The advancement of object detection research on point clouds has significantly increased the availability of datasets.
Within the context of autonomous driving, there are different
dataset such as KITTI \cite{KITTI}, Waymo Open \cite{Waymo} and nuScenes \cite{nuScenes}
and others which are available online, and include LiDAR as part of the dataset. These datasets
contain a collection of diverse urban scenes and dense position and class annotations.
These datasets provide us with valuable resources for validating autonomous driving algorithms, 
enabling the development of more accurate and robust perception systems.

Whilst these datasets are useful, in reality, it is impossible  to
test and improve autonomous driving systems in all potential 
real-world scenarios. Simulation of the perception system and the
physical world becomes an important approach for testing and
improving the safety of autonomous driving. Indeed, it was shown
\cite{abu2018augmented}\cite{ros2016synthia}\cite{richter2016playing}\cite{shafaei2016play}\cite{hanke2017generation} that simulation-based approaches can achieve state-of-the-art results when
autonomous driving models were trained on synthetic data.

LiDAR is one of the most important sensors for autonomous driving and many works have focused on the simulation of LiDAR. In \cite{fang2020augmented}\cite{fang2018simulating}, a LiDAR simulator which can augment real point clouds with virtual objects is implemented. The point cloud sweeps of real-world surroundings (usually to be used as a background) are generated from real LiDAR scanners mounted on top of a vehicle. The point cloud of a virtual object is obtained by scanning handcrafted CAD models with virtual LiDAR. These works demonstrate the importance of position and orientation of scanning objects for detection and segmentation. A LiDAR simulation approach described in \cite{manivasagam2020lidarsim} leverages real-world data to enhance the realism of the simulated scenes.
This approach results in more diversity and realism in the simulated scenes. It also describes a learning system that models the residual physics that are not captured by graphics rendering, which further boosts the realism of the simulation. 
For a drone usage scenario, the study in \cite{lidarsim} employs LiDAR simulation to evaluate the performance of detect-and-avoid systems for unmanned aerial systems (UAS).

In this work, we enhance the AI training dataset by augmenting real LiDAR data with additional synthetic point clouds generated using a Digital Twin. Unlike traditional Euclidean augmentation techniques that arbitrarily reposition objects, our approach ensures that inserted point clouds accurately reflect the LiDAR scanning pattern at their new locations. This physics-informed augmentation improves realism by preserving the sensor’s spatial and temporal characteristics, leading to more representative training data.

%% file: sec/method.tex
\section{Detection Model} \label{sec:methodology}

In this section, we propose an airborne PointPillars model that meets the requirements of scan space extension and the real-time processing needs of airborne applications.

\subsection{Standard PointPillars} \label{subsec:pointpillars}

PointPillars is an object detection algorithm, originally applied to automotive hazard perception on the ground. 
It can operate on 3D point clouds generated by LiDAR sensors. As shown in Fig. \ref{fig:pointpillar-arch}, the PointPillars network consists of three main components: the voxelization layer (pillar feature net), the backbone layer, and the detection head. The voxelization layer takes raw point cloud data as input and converts it into a sparse 3D voxel grid representation. The voxel grid is organized into pillars, which are vertical stacks of voxels that preserve the spatial and geometric information of the original point cloud data. The feature dimension of each point is denoted by $D$, encompassing nine dimensions: the spatial coordinates $x$, $y$ and $z$; reflectance $r$; the pillar center offset values $x_{p}$ and $y_{p}$; and the distance to the arithmetic mean of the pillar $x{_c}$, $y_{c}$ and $z_{c}$. 
Points within each pillar are first processed through a 1D convolution block.
The maximum values from each pillar are then taken as features and scattered to their corresponding positions on the 2D pseudo-image of the pillar grid. The backbone layer is a convolutional neural network (CNN) that takes as input the features of the points in each pillar and extracts high-level semantic features that capture the spatial and semantic characteristics of the objects in the point cloud data. 
The backbone layer is the most computationally expensive part, consuming 81.8\% of the total computation (see Section \ref{sec:real-time-performance}).
The detection head takes as input the features extracted by the backbone layer and generates a set of predicted 3D bounding boxes for each object in the point cloud data. 
The detection head predicts the objectness score for each voxel in the point cloud data and then refines the location and size of each predicted box.


\subsection{Airborne PointPillars} \label{subsec:adaptation}

In autonomous driving applications the LiDAR sensor is mounted on top of a vehicle and the point cloud data is collected from a ground-level perspective. In airborne sense and detect, the LiDAR sensor observes the full 3D space, with objects of interest frequently appearing above and below the horizontal plane of the drone-mounted LiDAR.
Our use case (maintenance inspection survey) involves flight at altitudes of up to 120m (due to regulations) but can also extend below the surrounding ground level, such as when descending into the Harburg Lock (see Section \ref{subsec:Hamburg}) to conduct inspections within its structural enclosure. In photogrammetry and high-resolution LiDAR surveys, flights often require drones to operate in close-proximity to structures to achieve the necessary ground sample density (GSD), sometimes within just a few meters of the target surface. This increases the complexity of the object detection challenge, as drones must reliably detect and avoid both static infrastructure and other airborne assets in constrained environments.
While the standard PointPillars was trained to detect multiple classes of object (cars, pedestrians, etc.), we focused on a single object class and annotated only observations of drones in the point cloud data. 

In the PointPillars architecture, the 3D space is divided into a set of pillars, and each pillar corresponds to a 2D slice of the 3D space. The anchors are predefined 3D bounding boxes that are used to classify and regress the objects within each pillar. By default, the PointPillars algorithm uses fixed-size anchor boxes and the z-center of the anchor boxes is usually fixed at a constant value, since the elevation of the object being detected is assumed to be relatively constant in the local area covered by the LiDAR sensor. 
PointPillars performs object localization by first regressing the object's position in the x-y plane, followed by additional regression steps to determine height and elevation. This means that if a dataset includes objects at varying altitudes and more anchor boxes are added with different elevations, the PointPillars model can be trained to predict the 3D locations of objects at varying altitudes.

To better detect airborne objects such as drones with varying elevation, we made the following adaptation to the standard PointPillars. We horizontally stratify the LiDAR field of view into layers where each layer extends to 1m in height (typically 10 layers above and 10 layers below the elevation of the LiDAR). We allocate an anchor box of size 1.6m×1.6m×1.0m (length × width × height) to each layer, with the center of the anchor box located at the center of each layer. We also assign different class names to each layer, from $drone\_0$, $drone\_1$, ..., $drone\_20$. Each class has a corresponding anchor box. The optimal elevation prediction will be related to both the z-center localization accuracy and the classification accuracy. As shown in the standard PointPillars, the localization regression residual along the z-axis is given by Equation \ref{equ: loss-z},  where $z^{gt}$ and $z^{a}$ are the z-center of ground truth box and anchor box respectively, $h^{a}$ is the height of the anchor box. The object classification loss is given by Equation \ref{equ: loss-class}, where $p^{a}$ is the class probability of an anchor, $\alpha$ and $\gamma$ are parameters used to adjust the loss function in order to address class imbalance,
\begin{equation}
\Delta z = \frac{z^{gt}-z^{a}}{h^{a}}
\label {equ: loss-z}
\end{equation}

\begin{equation}
L_{cls} = -\alpha_a(1-p^{a})^\gamma 
\label {equ: loss-class}
\end{equation}
 here $\alpha$ and $\gamma$ are set to constants of 0.25 and 2 respectively as in the standard PointPillars.

This approach allows the algorithm to better detect objects with varying elevation, such as a drone. The size and spacing of the layers are determined based on the characteristics of the drone being detected. We use the same loss functions introduced in the standard PointPillars. 

It is important to note that adding more layers along the z-axis may increase the computational cost of the algorithm. Therefore, it is important to balance the number of layers with the performance requirements of the application. A trade-off between accuracy and speed should be considered when selecting the number of layers.

To evaluate the performance, the model was tested during supervised flights at Newcastle Aerodrome, Ireland (see Appendix~\ref{app:newcastle}) and Hamburg Port, Germany  (see Appendix~\ref{app:hamburg}). The two test sites comprise a mix of built and natural environment and provide distinct background scenery from one another. The Newcastle site is a coastal railway bridge over a tidal inlet and is surrounded by rocks, gravel, and sand dunes. The Hamburg site is the Harburg Lock, which is a navigable waterway system that consists of a concrete chamber with gates at either end, surrounded by buildings and trees. Maintenance inspection surveys were performed at both sites involving multiple drones deployed with camera and LiDAR sensors to collect point and pixel datasets for structural analysis. The flight profiles at both sites were also significantly different, in particular the perspective and relative vantage points from which drones were observed in the LiDAR training data. This is significant as the detection model decorates input points to encode the shape, appearance, and relative position of observed objects, while the grid and pillar layout encodes the spatial relationship between the object and its surrounding environment.

\subsection{Optimization} \label{subsec:pointpillars}
We propose a new algorithm for sparse convolution and apply it to accelerate the backbone layer, which is the most computationally expensive part of the detection model.

\subsubsection{Sparse convolution with scatter operation} \label{subsec:sparseconv}

\begin{figure}[h]
\centering
   \begin{subfigure}[b]{0.48\textwidth}
    \includegraphics[width=\textwidth]{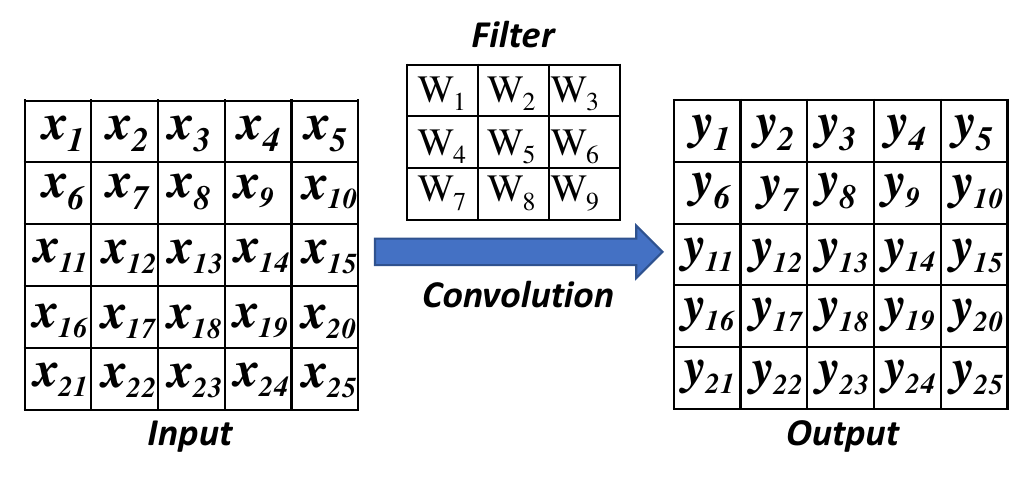}
    \vspace*{-1cm}
    \caption{}
   \label{fig:standardConv}
\end{subfigure}
\;
\begin{subfigure}[b]{0.5\textwidth}
    \includegraphics[width=\textwidth]{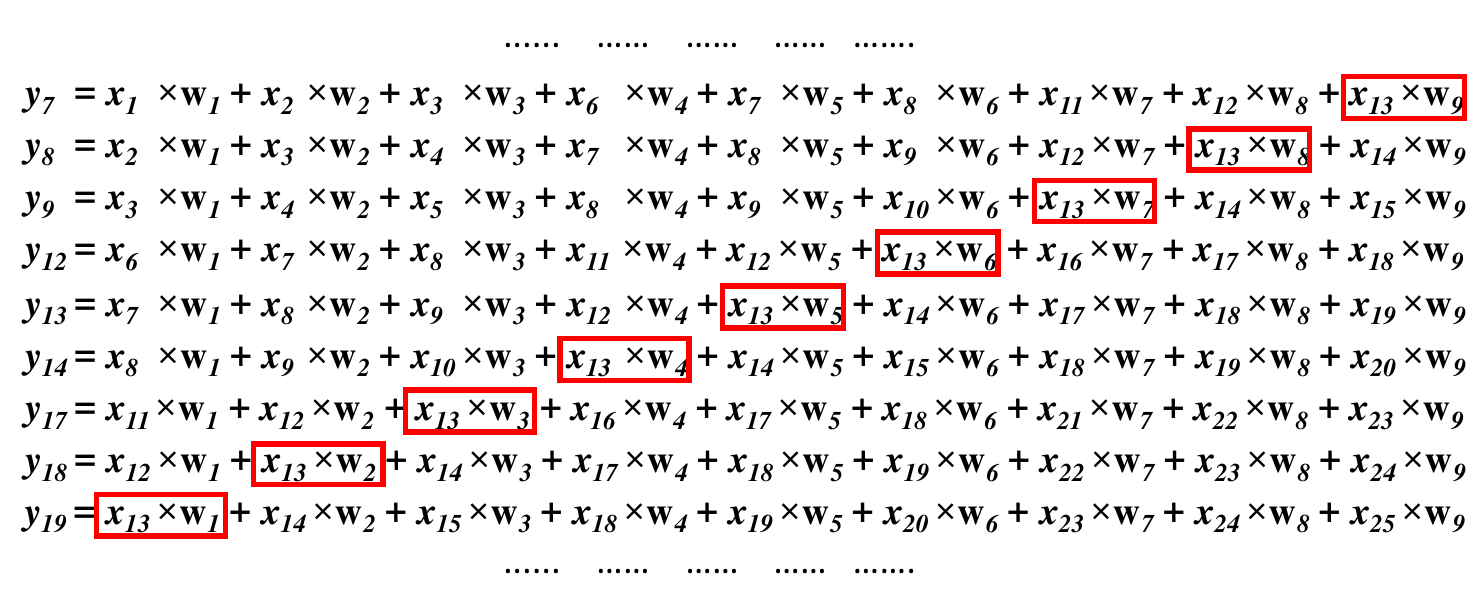}
    \caption{}
   \label{fig:eq}
\end{subfigure}
\;
\begin{subfigure}[b]{0.48\textwidth}
    \includegraphics[width=\textwidth]{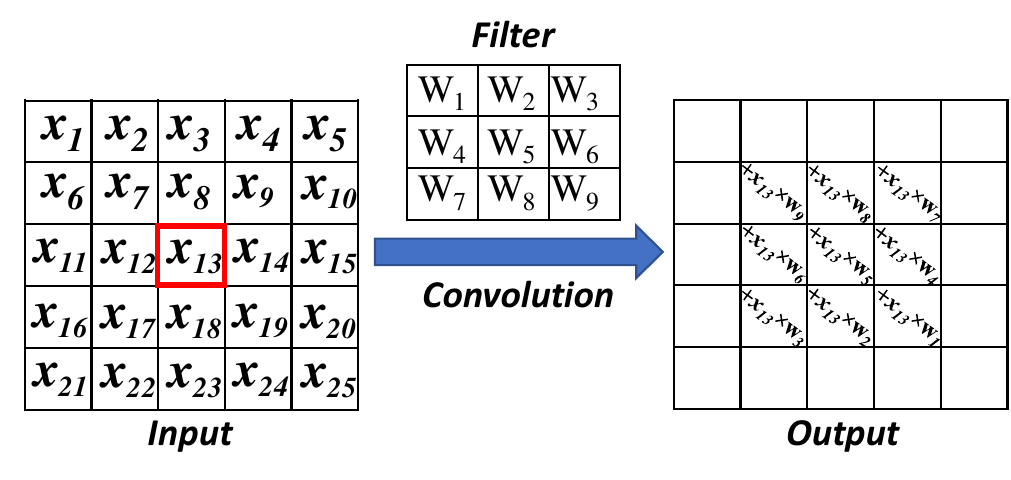}
    \vspace*{-1cm}
    \caption{}
   \label{fig:scatterConv}
\end{subfigure}
\caption{(a) Standard convolution where $x_i, w_i, y_i$ represent the input, filter and output, respectively. (b) Computation of each output $y_i$; It also shows how $x_{13}$ is involved in the computation of these output elements. (c) The proposed Convolution with Scatter Operation.}
\end{figure}

We describe the convolution with scatter operation first.
With standard convolution, at each output location, the product between each element of the filter and the input element it overlaps with is
computed and the results are summed to obtain the output in the current location, see Fig. \ref{fig:standardConv} and \ref{fig:eq}. However, in standard convolution, observe that the same input element multiplied by different filter weights contributes to different output elements and the location of those output elements neighbor each other. For example, in Fig.\ref{fig:eq}, the input element $x_{13}$ is multiplied by different filter weights and the results are added to different output locations. Specifically, $x_{13} \times w_9$ is added to the location of $y_7$, $x_{13} \times w_8$ is added to the location of $y_8$ and so on. We can implement the convolution in an alternative way: multiply each input element with different filter weights and add the result to each corresponding output location which is decided by the filter weight location, see Fig. \ref{fig:scatterConv}. We call this implementation Convolution with Scatter Operation (Unlike standard convolution in which each neighboring pixel is multiplied by the corresponding filter weight and the result is added to the final output position, in this proposed convolution, each pixel is multiplied by different filter weights and the results are added to different output
positions. Therefore we call it the Convolution with Scatter Operation). 


Given an input image $I$ with size of $p \times  q$, a filter $W$ with size of $k \times k$ and the output image $O$, 
the general formula of the convolution with scattering operation is given as follows: 
\begin{equation}
\begin{split}
O_{i - m +  \floor {k/2}, j - n + \floor {k /2}} \; += I_{i,j} \times W_{m, n} \\
m \in \{0, ..., (k-1)\}, n \in \{0, ..., (k-1)\} \\
i \in \{0, ..., (p-1)\}, j \in \{0, ..., (q-1)\}
\end{split}
\end{equation}

For the input image with $C$ channels, the convolution with scattering operation can be expressed as follows:
\begin{equation}
\begin{split}
O_{i - m +  \floor {k/2}, j - n + \floor {k /2}} \; += \sum_{c=0} ^{c=C-1}{I_{i,j,c} \times W_{m, n,c}} \\
m \in \{0, ..., (k-1)\}, n \in \{0, ..., (k-1)\} \\
i \in \{0, ..., (p-1)\}, j \in \{0, ..., (q-1)\}
\end{split}
\end{equation}

Obviously, the sparse convolution can be implemented by applying the scatter operation to each active site (non-zero site) only.


The pseudocode of the convolution with scatter operation is given in Algorithm \ref{algorithm1}. 
It is clear the number of multiplications and additions required is the same as standard convolution, it is $O(p \times q \times k^2)$.
For sparse convolution with scatter operation, the time complexity is $O(l \times k^2)$ where $l$ is the number of active sites and $l \ll p\times q$.
\begin{algorithm}
\SetAlgoLined
\KwIn{An image $I(i,j)$ of size $p \times q$ and a filter $W(m,n)$ of size $k \times k$}
\KwOut{A new image $O(x,y)$ after convolution}
\For{$i \leftarrow 0$ \KwTo $p-1$}{
  \For{$j \leftarrow 0$ \KwTo $q-1$}{
    \For{$m \leftarrow 0$ \KwTo $k-1$}{
      \For{$n \leftarrow 0$ \KwTo $k-1$}{
	$x \leftarrow i - m +  \floor {k/2}$\;
	$y \leftarrow  j - n + \floor {k /2}$\;
        $O(x,y) \leftarrow O(x,y) + I(i, j) \times W(m,n)$\;
      }
    }
  }
}
\Return{$O(x,y)$}\;
\caption{Convolution with scatter operation}
\label{algorithm1}
\end{algorithm}

\subsubsection{GPU implementation} \label{subsubsec:spconv-gpu}

The convolution operation can be efficiently implemented on GPU using matrix multiplication, as shown by \cite{cudnn-arxiv}. This approach requires transforming the input image and the convolution kernel into matrices that can exploit the fast multiplication capabilities of the GPU. This transformation involves duplicating the input data to lower its dimensionality. 

In our GPU implementation of Sparse Convolution with Scatter Operation, we use matrix multiplication to perform the channel-wise multiplication and summation of the input and a filter kernel. To achieve a higher performance, we first use the GPU prefix sum  \cite{thrust} to compress the active sites in the input image into a dense format. Positions of all active sites are recorded in a 2D mask array, where an element of the 2D array is set to 1 if the corresponding site is an active site, otherwise it is set to 0. The prefix sum will then be applied to this 2D mask array.  
NVIDIA cuBLAS library  \cite{cublas}, a library of highly optimized matrix multiplication routines, is used to perform the channel-wise multiplication and summation. The results are then scattered to the corresponding output positions according to the scattering rule. To avoid write conflicts, we use the atomicAdd() function \cite{cuda} to perform the final addition. Unlike the method in \cite{cudnn-arxiv}, we do not apply any data lowering in our GPU implementation. Additionally, there is no rule book or hash table used in our GPU implementation, as the use of rule book or hash table increases the memory overhead and the computational complexity.

In this work, we present the initial implementation of the Sparse Convolution with Scatter Operation. Our GPU implementation follows the gather-GEMM-scatter dataflow \cite{torchsparse}, which needs to read and write DRAM at least three times, respectively. There is significant room for optimization using the following techniques: 1) perform computation on on-chip SRAM (shared memory) using tiling, which reduces access to DRAM memory, and 2) overlap memory access with computation. 

%% file: sec/training.tex
\section {Model Training} \label{sec:train}

Our training and testing workflows incorporate the requirements of the Specific Operation Risk Assessment \cite{SORA}. The methodology involves 5 steps:
\begin{enumerate}
  \item The operational site is mapped by a drone mounted LiDAR. The LiDAR is oriented in forward looking mode to capture point cloud data for the monitoring scenario.  During the mapping flights, multiple drones are flown in controlled patterns within the segregated airspace and under supervision. The observations of these drones in the LiDAR point cloud data are labelled using the recorded position and navigation data of each drone in the swarm to generate an annotated training dataset. \textcolor{black}{With the help of GNSS correction techniques such as Real Time Kinematics (RTK) and Post Processed Kinematics (PPK), we achieve a high-quality annotation in which the error can be controlled $<5cm$.}
  \item The annotated point data is augmented by inserting additional synthetic LiDAR data of drones in simulated safety critical scenarios.
  \item The blended reality dataset is used to train the model.
  \item Model performance is measured using real LiDAR data. Identified shortcomings are used to refine the training and model parameters and define strategic mitigations during mission planning, for example exclusion zones that might be challenging the model accuracy.
  \item The trained model is embedded on the drone and deployed for tactical monitoring of drones within the operating theater.
\end{enumerate}

\subsection {Training Data Collection and Labelling} \label{sec:realdata}

Collection of the training and testing datasets utilized two different LiDAR systems. Server-based training and lab testing used data collected with a DJI L1 LiDAR \cite{DJI-L1-LiDAR}, while the Livox Avia LiDAR \cite{Avia} is used to test the real-time embedded model on the drone. The systems are interchangeable as the L1 package contains a Livox Avia module, whilst incorporating an IMU and an active gimbal to compensate for motion disturbance and enable accurate recording of point data in absolute coordinates. However, the L1 does not support real-time data access and thus can not be used in sense and detect modes of operation.

\begin{figure}[ht]
\centering
\includegraphics[scale=0.55]{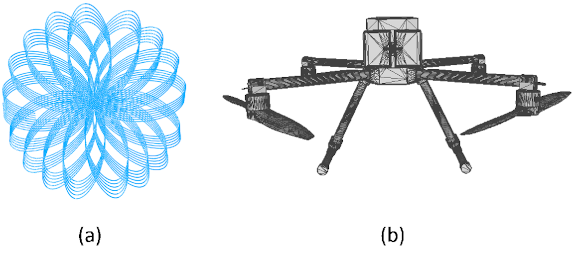}
 \caption{(a)  Livox Avia non-repetitive scanning pattern (100ms frame, 24k points) (b) The wireframe mesh model of the M300, which is defined by 207409 triangular facets.}%
\label{fig:scan-and-mesh}
\end{figure}

The Avia generates 240k points per second with two options for scanning mode. 
The repetitive line scanning is 70.4° (horizontal) × 4.5° (vertical) and is best suited for mapping in downward looking configuration while the drone moves forward. The alternative non-repetitive scanning pattern (see Fig.~\ref{fig:scan-and-mesh}(a)) is 70.4° (horizontal) × 77.2° (vertical) and is preferable for monitoring airspace.

\begin{figure}[ht]
\centering
\includegraphics[scale=0.27]{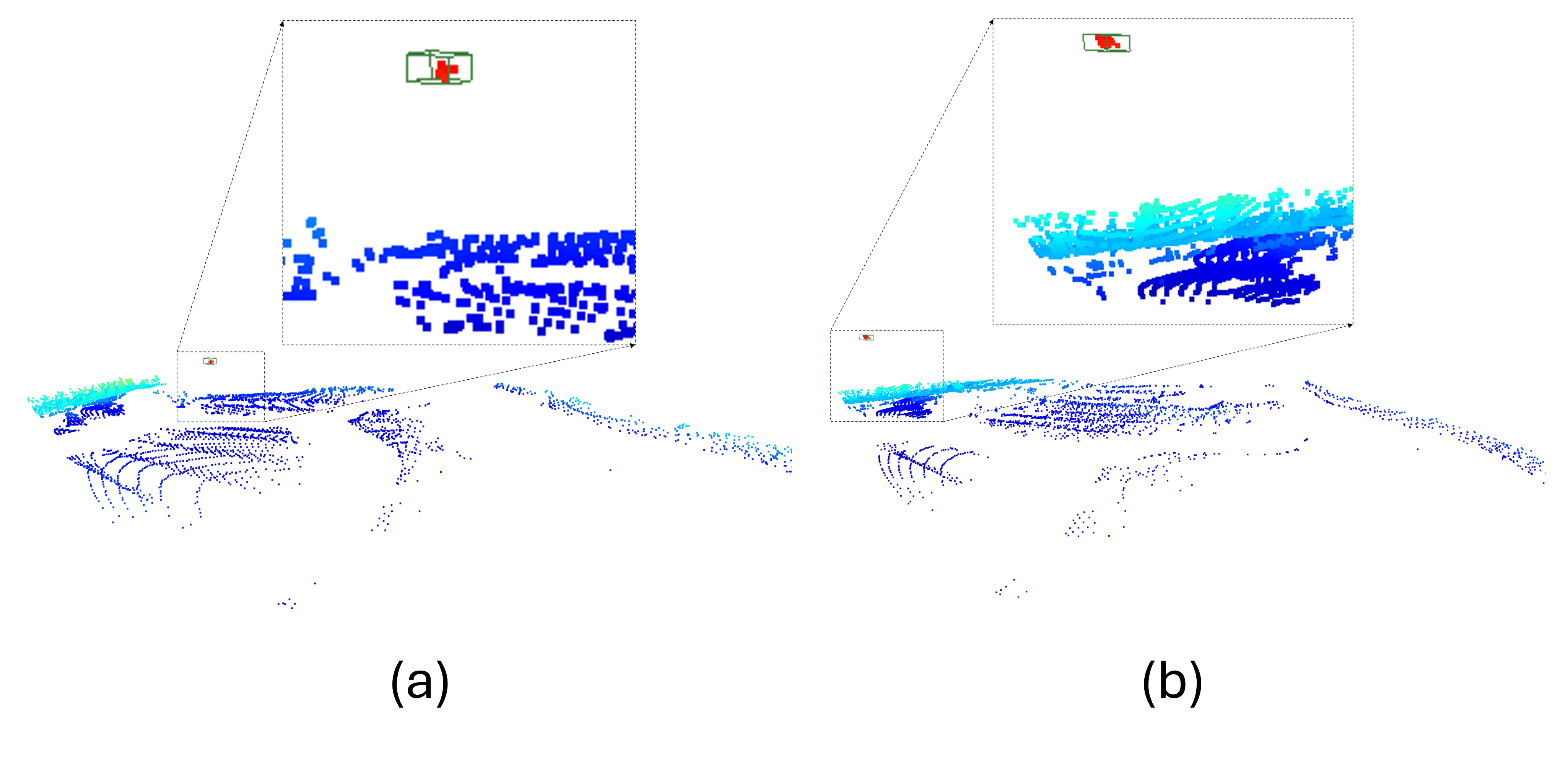}
\vspace{0.3cm}
\caption{A lateral view of a 100ms frame of LiDAR data, showing a swarm drone labeled with a green box. A portion of the frame containing the drone is enlarged for clarity. (a) The distance between the LiDAR and the swarm drone is  13.51m. (b) The distance between the LiDAR and the swarm drone is  46.58m.}
\label{fig:side-view}
\end{figure}

L1 data exchange is supported by converting the raw LiDAR data into the interoperable LAS format (v1.2) and Smoothed Best Estimate (SBET) files for 3D reconstruction in the EGM96 geoid. The observations of swarm drones can then be labelled with bounding boxes calculated from the recorded flight logs. For each point, the LAS (Point Record Format 3) records the relative coordinate X, Y, Z and reflectivity along with the GNSS timestamp at microsecond precision. Leveraging RTK and PPK techniques, it ensures a positional accuracy within $5cm$.
The SBET file provides the timestamped 6 degree-of-freedom LiDAR pose information throughout the flight enabling transformation of the LAS point data between body, earth, and PointPillars coordinate systems. In the case of the real-time stream from the Avia, the data is already in body fixed frame aligned to the PointPillars coordinate system.  Example frames (100ms frame) with swarm drone labelled are shown in Fig. \ref{fig:side-view}. 

\subsection{Data Augmentation} \label{subsec:dataaug}

As described in Section \ref{sec:methodology}, 
the robustness of the detection model relies on the geometric and spatial variation in the point cloud. However, obtaining point clouds from every position of the field of view of a LiDAR sensor is, in reality, not feasible.
To address this limitation, training data augmentation is commonly employed.

Conventional data augmentation strategies expand the training dataset by applying
fixed transformations, such as rotation, scaling, and translation, to the original labeled objects. 
However, 
due to the 3D spatial characteristics of point clouds, it is possible that the LiDAR may never generate an augmented point cloud from its viewpoint in the physical world.
Especially when the LiDAR employs a non-uniform scan pattern (non-repetitive pattern).
The LiDAR never generates a point cloud that matches the rotation-augmented or translation-augmented points from its view point.
Also, unlike camera images, the scale of a 3D point cloud remains constant regardless of the distance between the LiDAR sensor and the object being scanned.
This potentially results in learned features from augmented data that contribute less or even negatively to the model's overall performance.
For a drone usage scenario, the study in \cite{lidarsim} employs LiDAR simulation to evaluate the performance of detect-and-avoid systems for unmanned aerial systems (UAS).

\begin{figure}[ht]
\centering
\includegraphics[scale=0.38]{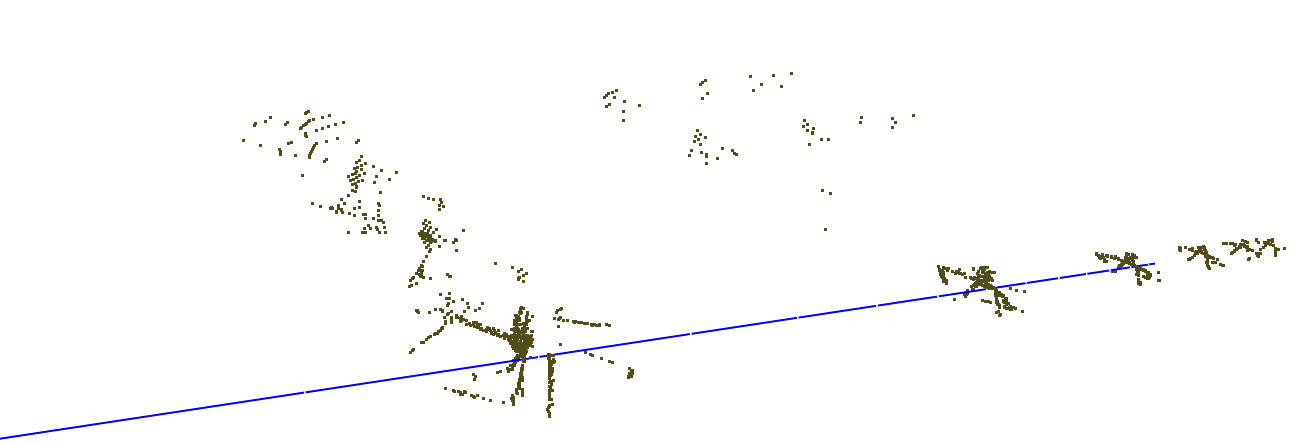}
\vspace{0.3cm}
\caption{Synthetic clusters of drones generated by ray tracing.}
\label{fig:synthetic-drone}
\end{figure}
\begin{figure}[ht]
    \centering
    \subfloat[\centering ]{{\includegraphics[width=7.2cm]{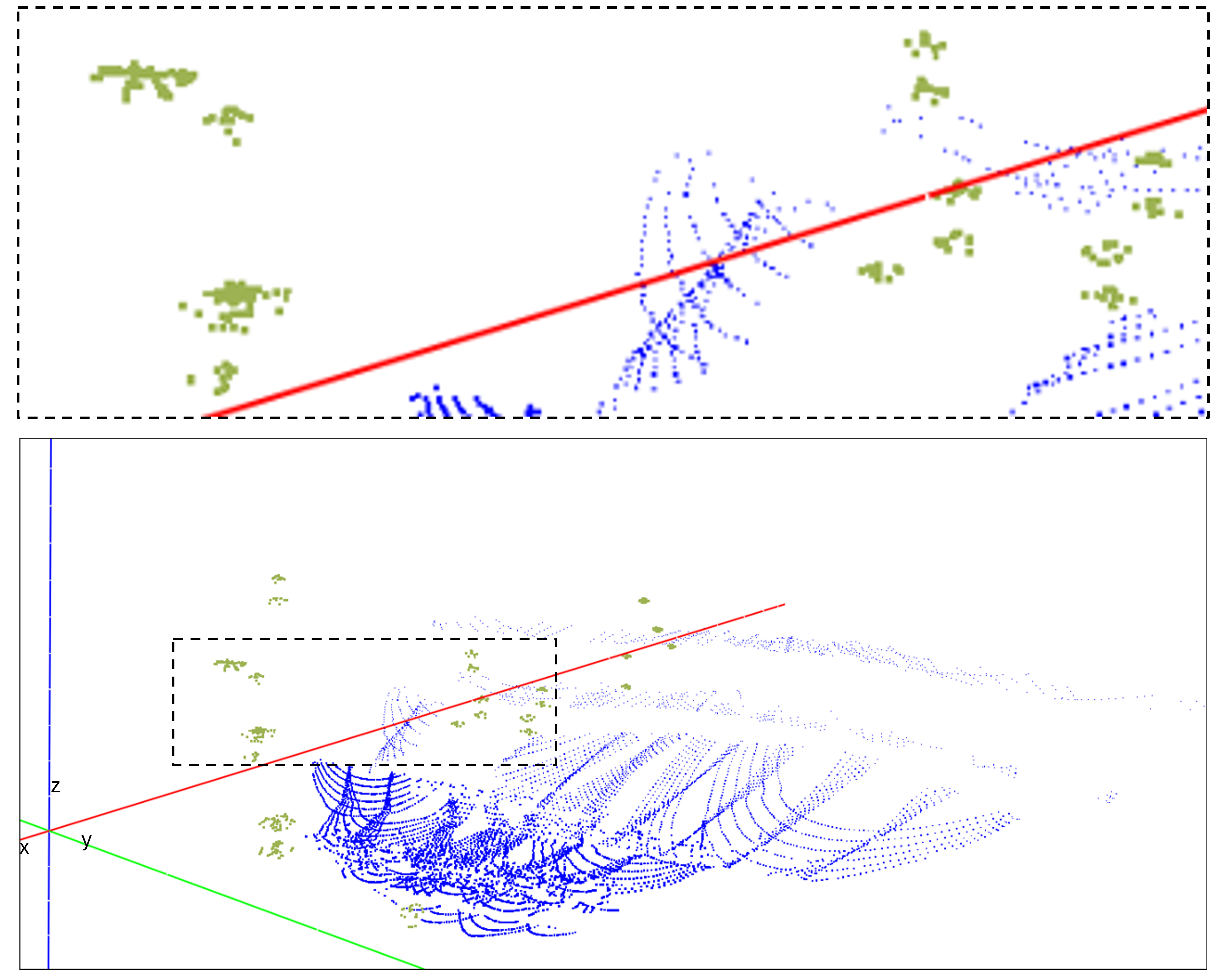} }}%
    \vspace{0.4cm}
    \qquad
    \subfloat[\centering ]{{\includegraphics[width=7.2cm]{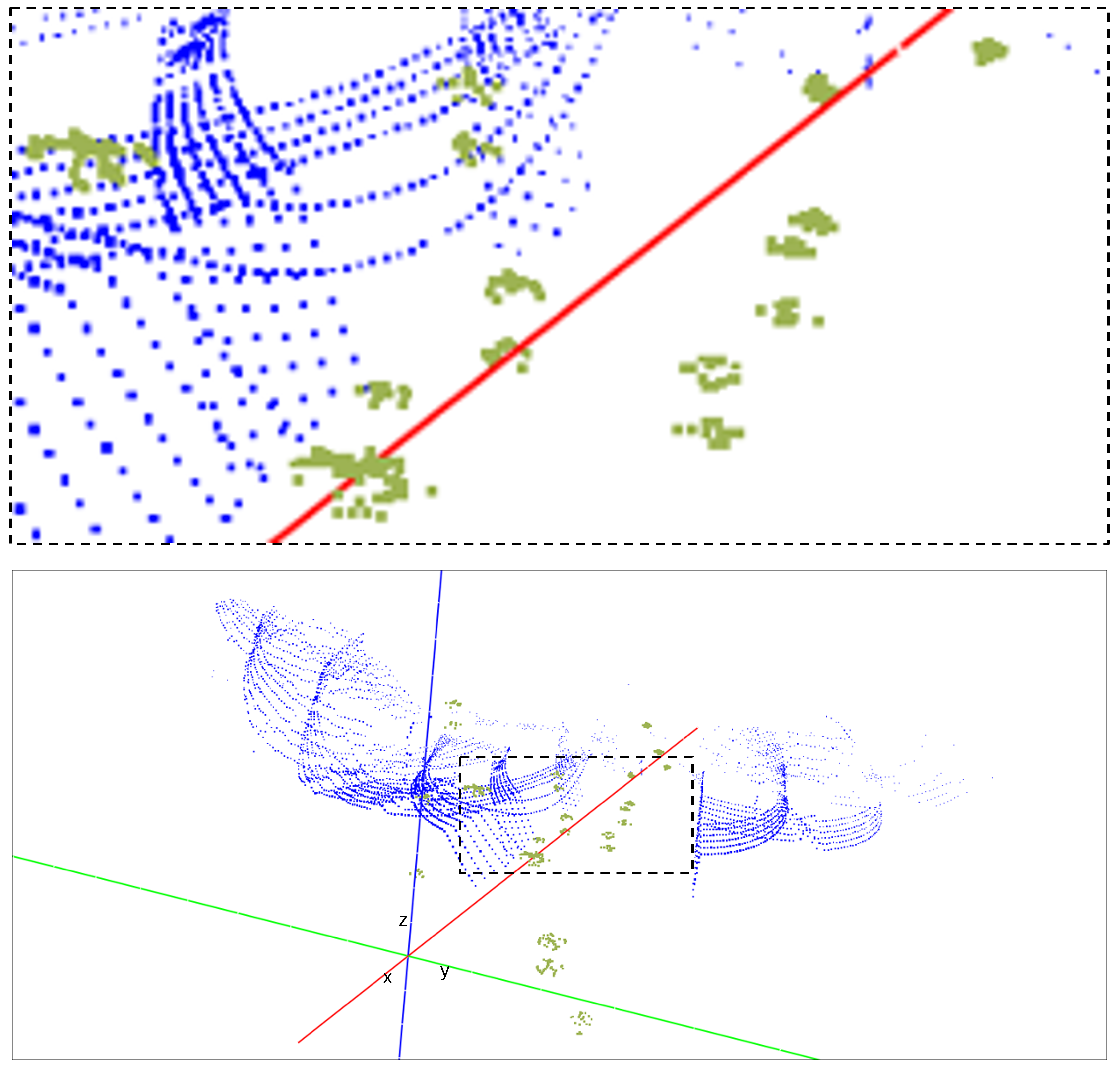} }}%
    \vspace{0.4cm}
   \caption{Visualization of simulated drones (represented as green points) inserted into two different real backgrounds. Each image is divided into two sections: the lower part provides a distant view of the entire scene, while the upper part shows an enlarged view of a selected area.}
    \label{fig:sim-inserted}%
\end{figure}

Our augmentation strategy instead uses a scenario Digital Twin to accurately model the relationship between the relative positions of the LiDAR and observed drones and the heterogeneous spatial and temporal resolution of the LiDAR sensor. We incorporate the NVIDIA OptiX Ray Tracing engine \cite{Optix} to accelerate the simulation of the LiDAR, for details see Appendix~\ref{appendix:sim}.
The simulated points generated by the Digital Twin exhibit a sampling pattern and sensing distortion that matches its relative location within the LiDAR field of view. Fig. \ref{fig:synthetic-drone} shows the simulated drone cluster generated by ray tracing a triangulated model of the M300 drone at different locations within the LiDAR field of view. Simulation of 100ms frames takes an average 0.2075ms on an 8 GPU server. This involves ray tracing a drone model with 207,409 triangles. In effect the simulator can generate data at 481 times the productivity of the real LiDAR. Analysis of the approach is presented in the results section. Fig. \ref{fig:sim-inserted} shows some examples of simulated drones which are inserted into a real background. The number of points of each drone is greater than 10 and all drone positions are randomly selected.
The Digital Twin enables the development cycle to virtualize the cost and risk of collecting data in safety critical scenarios.

\subsection{Training Parameter} \label{subsec:traingingpara}

The parameters for the training process are as follows.
For a single GPU, we limit the batch size to 4 due to memory constraints. However, when training with 8 GPUs, we increase the batch size to 32. During our experiments, we noticed that after 128 epochs, the accuracy did not improve further. Regarding the learning rate, we deviate from the standard PointPillars paper (which used 0.0001) and set it to 0.001. Our empirical observations indicate that this value is more suitable for our use case. Safety is critical in our use case scenario, so we avoid any point cloud sampling. Otherwise, small objects like drones might be excluded from the scene. In our use case, we have not encountered any situations where sampling out occurs when we set the maximum number of points each pillar to 100 and the maximum number of pillars to 12,000. However, these two values are changeable, dependent on the use case. To determine the optimal threshold values for positive IoU and negative IoU, we conducted numerous experiments with various combinations. Our observations indicate that the combination of 0.4 for positive IoU and 0.35 for negative IoU yields the highest accuracy.


%% file: sec/testing.tex
\section{Model Test}  \label{TP-FN-F1}

\subsection{Model Test with Digital Twin} \label{subsec:directivity-response}

\begin{figure}[ht]
\centering
\includegraphics[scale=0.375]{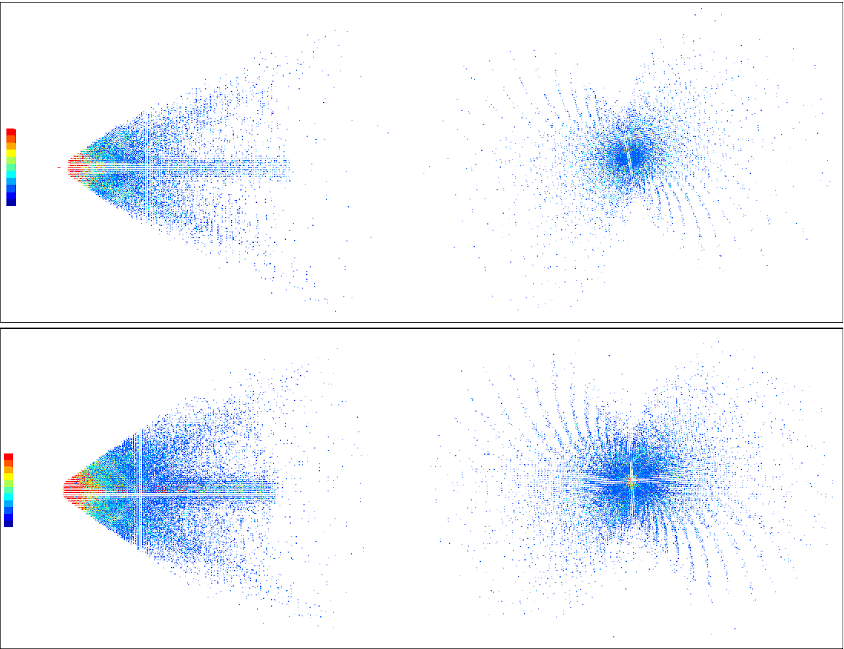}
\caption{(top) 100ms scanning time and 4 points per voxel threshold, and (bottom) 200ms scanning time. Each square block in the heat map represents the number of points per voxel threshold. The deep blue box in the bottom indicates that the minimum number of points per voxel is 1. The red box in the top indicates the minimum number of points per voxel is 20.}
\label{fig:pc4}
\end{figure}

\begin{figure}[ht]
\centering
\includegraphics[scale=0.375]{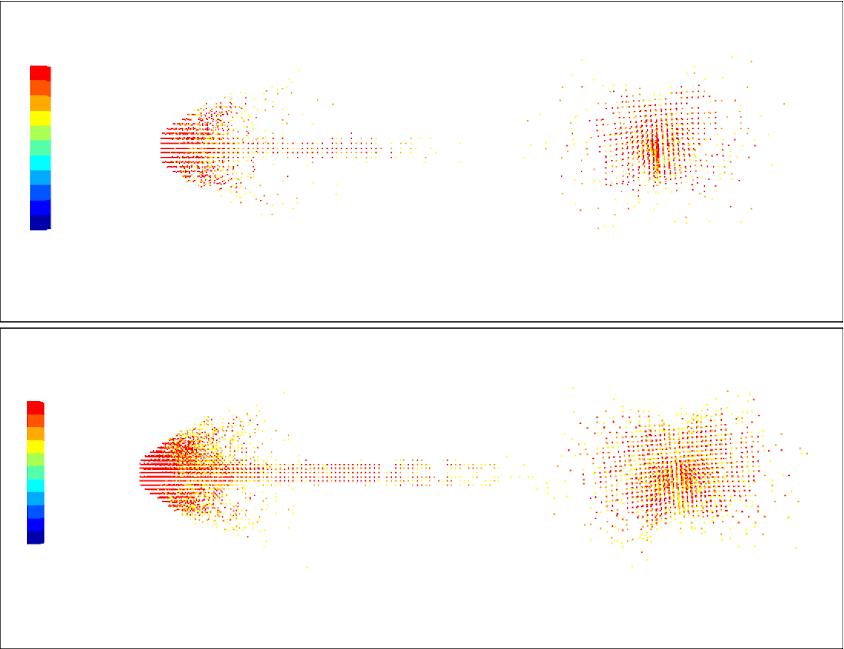}
\caption{(top) 100ms scanning time and 14 points per voxel threshold, and (bottom) 200ms scanning time and 14 points per voxel threshold.}
\label{fig:pc14}
\end{figure}

The LiDAR point data is accumulated into time windowed frames and then input to the detection model. The optimal time window depends on several factors. The upper threshold is determined by the acceptable computational latency which increases in proportion to the number of points in each frame, while the lower threshold is determined by the accuracy of the DNN.
To assess the detection model's performance across different datasets and parameters, we utilize high-fidelity drone and LiDAR simulations within a scenario-based Digital Twin. 
The selected time window affects the number of points that can be returned from an object within the LiDAR field of view. The heterogeneous LiDAR scanning pattern generates a non-uniform directivity response. This affects the object detection performance, with significant variation in the number of laser-returns backscattered by objects observed at different locations within the LiDAR field of view. The directivity response is a key factor in pre-mission risk assessment as drone observations with sparse point clouds will register as false negatives. 
Fig.~\ref{fig:pc4} and \ref{fig:pc14} illustrate the directivity pattern of LiDAR scanning with different time windows and sensing thresholds. The field of view of the LiDAR is discretized into cubic voxels of 1.0m side length. 
The drone mesh model (see Fig.~\ref{fig:scan-and-mesh}(b)) is placed at centre of each voxel. 
These voxels are then ray-traced over a time-windowed interval. Voxels that do not achieve the required number of ray intersections are excluded from the visualization. Only the center points of voxels that surpass the threshold are visualized  and colored based on the number of intersections, i.e., generated points. 

The detection model is tested within the digital twin using different point cloud data obtained from the scenario site with different configurations and varying frame time windows. 
Our experiments demonstrate that 100ms is an acceptable trade-off between latency and the number of points that are sent to the inference model for analysis. 

\subsection{Metrics and Error Budget for Model Test} \label{subsec:metric-error-budget}

To evaluate the performance of the trained model we use the recall, precision, and F1 scores. The recall, precision, and F1 score are calculated using IoU (Intersection over Union) scores. The definitions of each metric can be found in Appendix~\ref{appendix_metric}.  

In the sense and detect use case it is essential to minimize the number of false negatives to ensure the safety of the system and reduce risk of accident due to collision. False negatives mean that an airborne hazard is not detected, leading to a potential mid-air collision.  Conversely, false positives occur when the system detects an object or obstacle that is not actually present, leading to unnecessary actions or alerts. 
Minimizing false negatives is important because failing to detect real threats can lead to severe consequences, such as crash, rather than just minor inconveniences.

We use the following thresholds for IoU measurements:
\begin{itemize}
	    \item True Positive: at least 30$\%$ IoU between ground truth and inference bounding boxes.
	    \item False Negative: undetected ground-truth bounding box, i.e. the frame contains a drone.
	    \item False Positive: misplaced detection, i.e. a bounding box is placed in an incorrect location.
\end{itemize}

The IoU criterion of at least 30$\%$ between the inference label and ground truth label is justified by the relative volume of the drone compared within the overall scene. For example, the standard PointPillars model validated on KITTI uses a 50$\%$ threshold for pedestrians. In addition, incorporating the error budget of the positional estimation extends the volume of the ground truth bounding box label significantly. The error budget of the drone positional estimation can vary from 2-3cm for RTK and PPK solutions up to 10s of meters. For example, an RTK system in convergence mode degrades to 20-30cm accuracy, while if the RTK drops out at any stage, the whole flight reverts to 2-3m DGPS accuracy. As the drone occupies a 1.6m×1.6m×1.0m cube, this challenges ground truth labelling and biases IoU measurements during training, validation, and testing. 


While the drones can be required to fly a path that risks denial of GNSS and communications due to multi-path and signal occlusion by large civil engineering infrastructure, a tracking-by-detection algorithm is incorporated into the labelling. 
Tracking has been incorporated into point cloud-based machine learning workflows previously as in \cite{wang2019latte}\cite{sane}. We utilize a simplified approach of \cite{wang2019latte}, who employ a Kalman filter to predict the updated location and yaw of rigid bounding boxes for objects such as cars. In our approach, we simply re-center the bounding box from the previous frame around the enclosed points within the new frame and estimate the yaw angle. 
The algorithm begins with an initial set of detections performed by the deep learning-based detection algorithm. These detections correspond to objects within the scene and serve as the starting point for subsequent tracking. Bounding boxes are associated with these detected objects, encapsulating each object of interest. The approximate speed of the monitoring drone which is equipped with LiDAR, is estimated by analyzing the positional change between two consecutive detections. In each successive frame, the position of tracked objects is estimated by re-centering the bounding box to the centroid of a sparse point cluster, which represents the object's current location. If there are no points inside the re-centered bounding box, the frame is skipped, and the algorithm proceeds to the next frame. The speed information is used for the re-centering process. In practice, the deep learning-based detection algorithm and tracking algorithm are dynamically used. If the deep learning-based detection successfully identifies objects in the current frame, we proceed to the next frame. Otherwise, we switch to using a tracking algorithm for that frame. The speed estimation of the monitoring drone is updated using the latest two consecutive detections. The integration of tracking with detection aims to minimize both false positives (incorrectly identified objects) and false negatives (missed detections). The results of the strategy are analyzed in the results section.

%% file: sec/experimental-system.tex
\section{Hardware design and system integration}\label{systemArch}
\subsection{System Design Requirement}
For real-time sense and detect of drones in a safety scenario,
the system design must adhere to the following requirements:
\setlist{nosep}
\begin{itemize}
\item Latency: The frame-to-frame processing time must be minimized to ensure timely responses in dynamic environments.
\item	 Range: The sensor should provide sufficient coverage for reliable detection across the required area.
\item	 Accuracy: The system should achieve high accuracy to ensure that sensor provides reliable detection.
\item	 Independent of external aiding systems: The system should operate effectively without relying on external aids like GNSS. For example, GNSS can introduce errors such as GPS drift \cite{GPSdrift}.
\item Localization: The system must support 3D spatial localization to meet the specific requirements of the use case. 
\item Weight and Power Limitations: The design must accommodate the constraints of the drone's payload capacity and battery power.
\vspace{0.2cm}
\end{itemize}

\subsection{System Integration}
\begin{figure}[ht]
\centering
\includegraphics[scale=0.55]{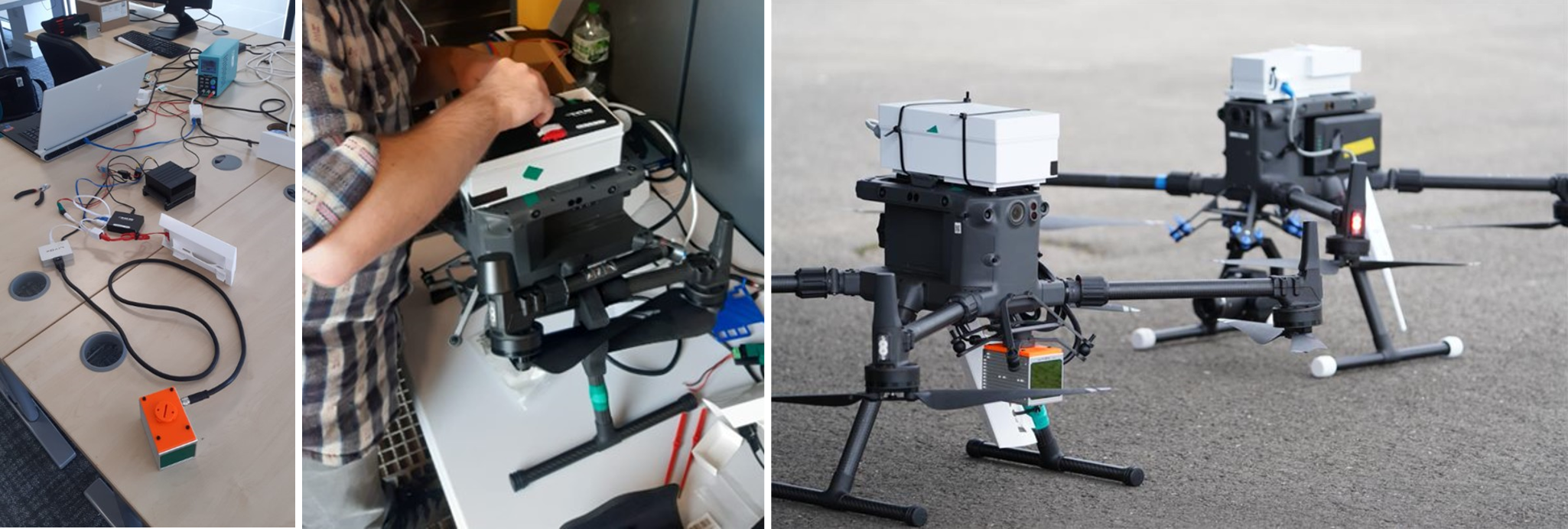}
\caption{Our real-time embedded system for object detection.}
\label{fig:real-time-system}
\end{figure}

The inference model is embedded on an Nvidia AGX mounted on a DJI M300 drone. The DJI M300 provides a power supply of 96W at 24VDC for the payload and has a 2.7kg payload lift capacity. Our integrated system always weighs less than this limit. The drone's battery provides sufficient power to all hardware components.
The AGX and peripherals are enclosed in a weatherproof housing and mounted with 3D printed brackets as shown in Fig. \ref{fig:real-time-system}. Aside from the PSU requirement, the bolt on system is agnostic of the host drone platform. A collection of power converters provide the required step up/down voltages between the XT30 24VDC supply and the payload electronic components.

\begin{figure}[ht]
\centering
\includegraphics[scale=0.27]{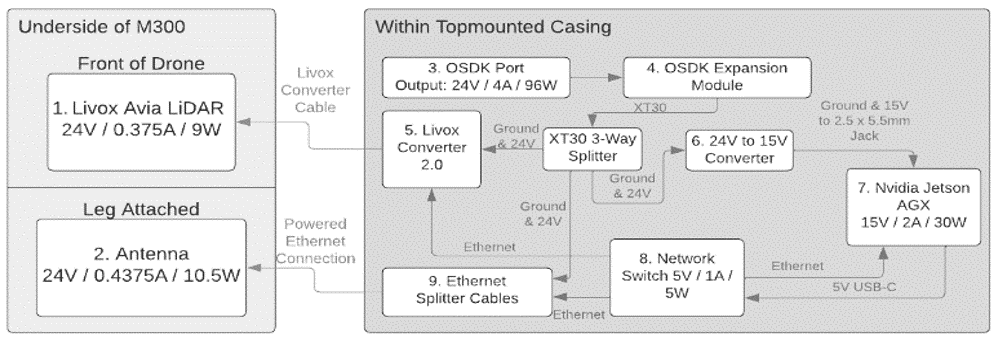}
\caption{Illustration of the real-time system schematic.}
\label{fig:system-schematic}
\end{figure}

The perception sensor is a Livox Avia LiDAR and it is connected to the AGX computer via a network hub. The network hub provides a bi-directional communications link to the ground via a Wi-Fi antenna mounted on the drone chassis. This enables the software running on the AGX to send alert messages to the pilot on the ground over a TCP-IP socket. An XRDP remote desktop protocol agent enables the operator to configure the onboard components during pre-flight checks. Before take-off the XRDP is disconnected to minimize bandwidth and GPU load and the TCP-IP network socket is launched to provide alerts to ground. The system schematic is shown in Fig. \ref{fig:system-schematic}.

While a denser point clouds can boost detection accuracy, low latency is essential for real-time hazard response.
According to \cite{aviaLatancy}, 
the Avia LiDAR is capable of covering most of the field of view (FOV) in 0.1 seconds (100 milliseconds). 
On the other hand, our experiments demonstrate that a latency of 100ms achieves an effective balance between two competing factors: ensuring sufficient accuracy and delivering a timely response for safety. Through experimentation, we determine that the system's maximum detection range is roughly 68m. To comprehensively assess the accuracy of our drone detection system in safety-critical scenarios, our evaluation considers three key metrics: precision, recall, and F1-score. 

%% file: sec/experimental-result.tex
\section{Performance of Sparse Convolution with Scatter Operation} \label{sec:spconv-gpu-performance}

\begin{table*}[th]
\centering
\caption{Comparison between our model and PCDet models.}
\begin{tabular}{|c|c|c|c|}
  \hline
 \textbf{3D Detection Model} &   \textbf{Modification} &  \textbf{Precision (\%)} &  \textbf{Inference time (ms)} \\
  \hline
 Centerpoint-PillarNet & Extend Z-axis range & 89.38 & 34.13\\
  \hline
DSVT-PillarNet & Extend Z-axis range & 67.86 & 62.25\\
  \hline
 \multirow{2}{*}{Our model (modified PointPillars)} & Extend Z-axis range only & 83.53 &14.16\\
\cline{2-4}
& Extend Z-axis range + our (sparse+submanifold)  convolution & 82.60 &6.13\\
 \hline
\end{tabular}
\label{tab:comp-other-model}
\end{table*}

\begin{figure}[tbp]
\centering
  \begin{subfigure}[t]{0.23\textwidth}
  \centering
    \includegraphics[width=\textwidth,  height=0.7\textwidth]{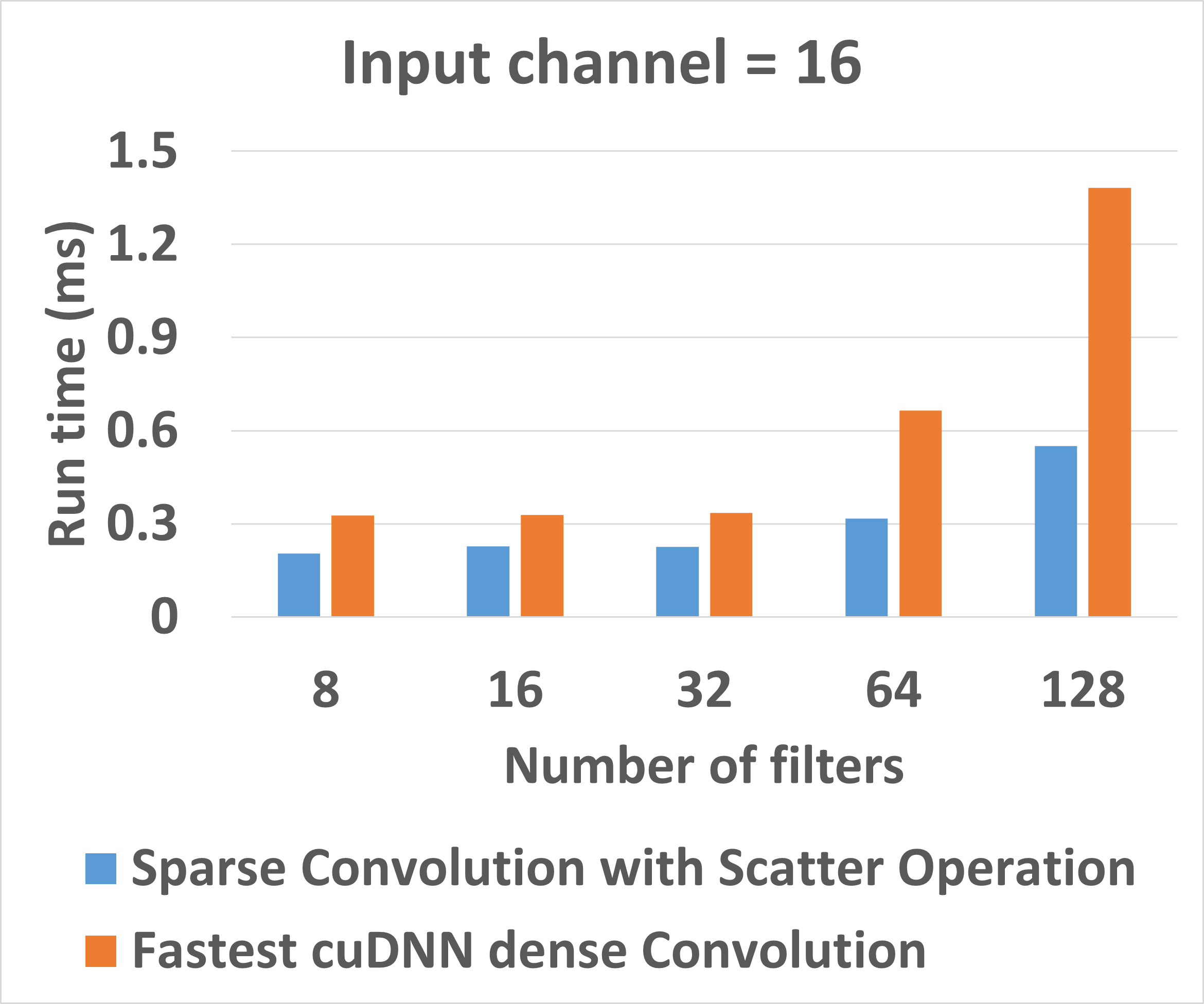}
    \caption{}
    \label{fig:performance1}
  \end{subfigure}
  \;
  \begin{subfigure}[t]{0.23\textwidth}
  \centering
    \includegraphics[width=\textwidth,  height=0.7\textwidth]{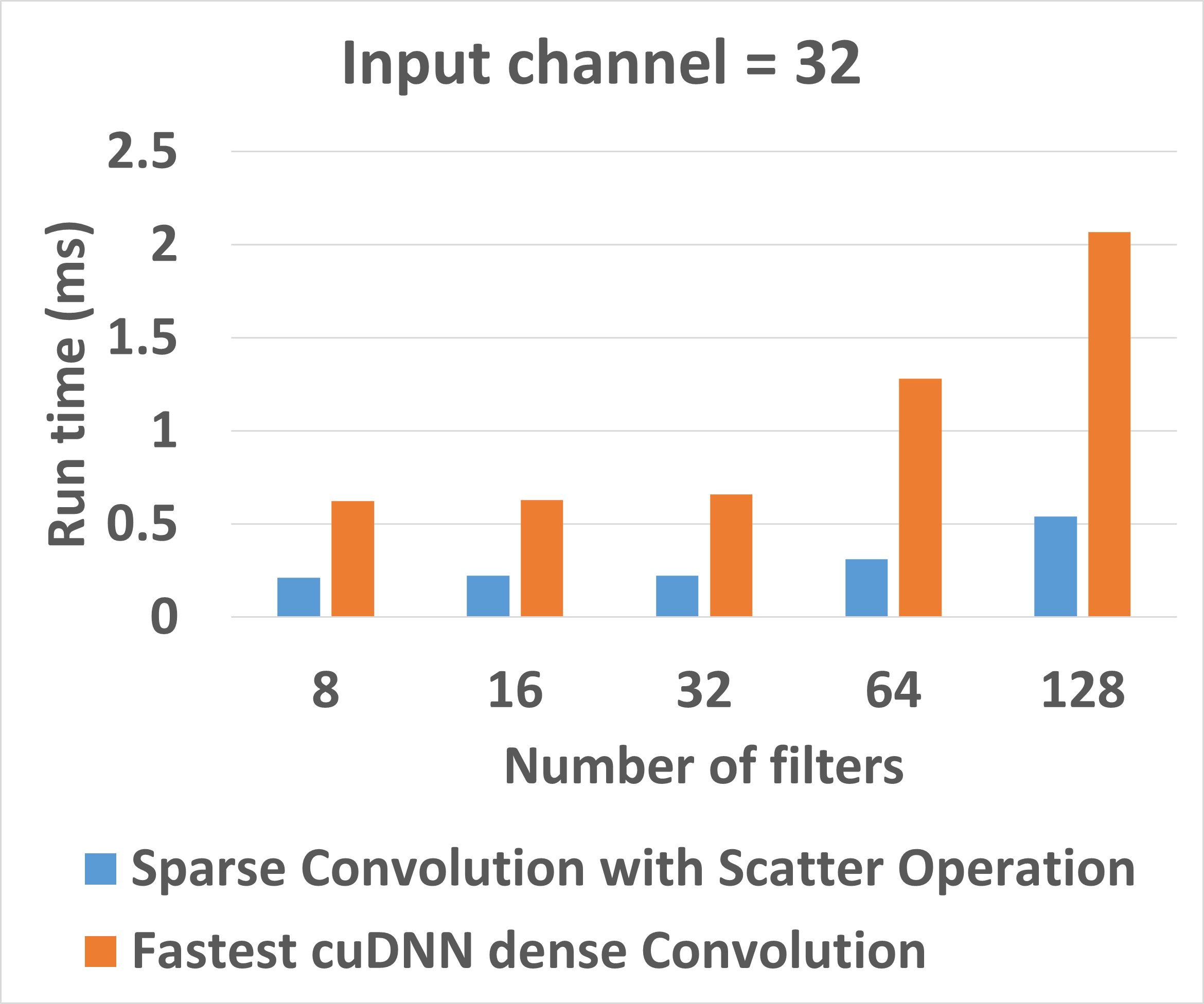}
    \caption{}
    \label{fig:performance2}
  \end{subfigure}
   \;

  \begin{subfigure}[t]{0.23\textwidth}
  \centering
    \includegraphics[width=\textwidth,  height=0.7\textwidth ]{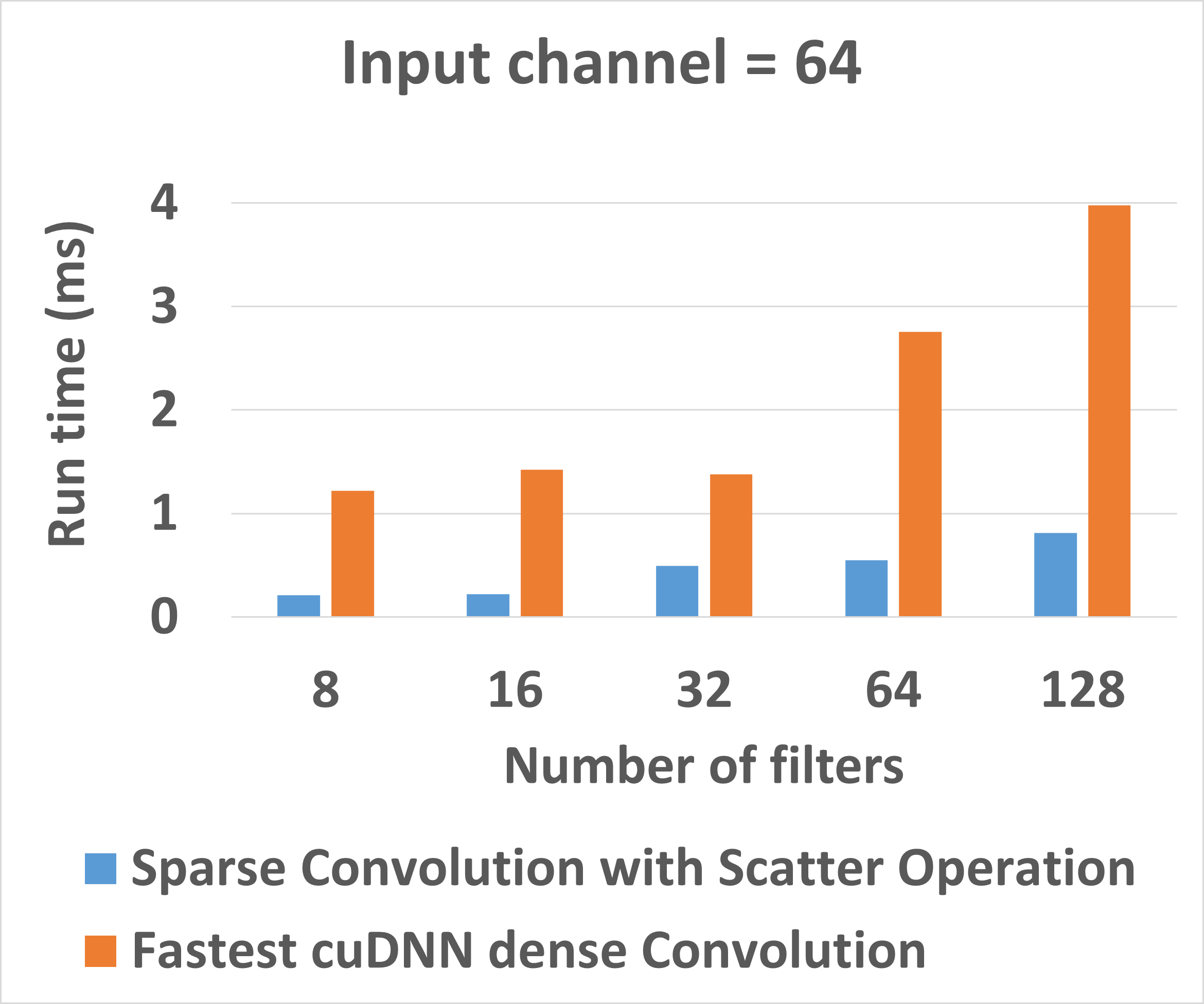}
    \caption{}
    \label{fig:performance3}
  \end{subfigure}
   \;
  \begin{subfigure}[t]{0.23\textwidth}
  \centering
    \includegraphics[width=\textwidth,  height=0.7\textwidth]{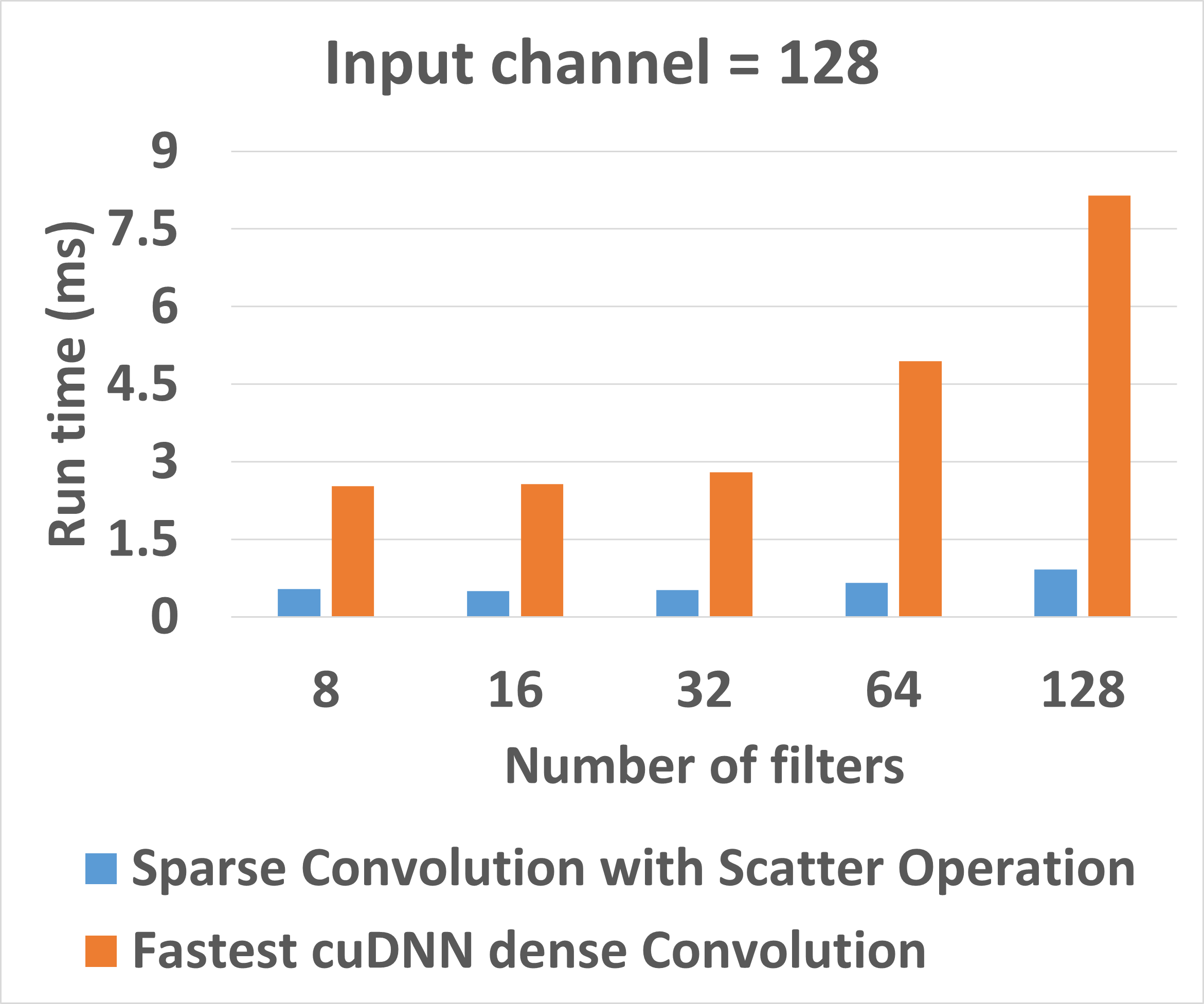}
    \caption{}
    \label{fig:performance4}
  \end{subfigure}
 \;
\caption{Performance of convolutions with varying parameters, evaluated over 1,000 images, with an average of 5,124 active sites per image; Size of image = $504 \times 504$, filter size = $3 \times 3$.}
\label{fig:performanceComparison}
\end{figure}

We use different experiments to evaluate the proposed sparse convolution with scatter operation. An NVIDIA Jetson Xavier (32GB SSD) and JetPack 5.0.2  with CUDA 11.4 is used for the performance evaluation.
The standard PointPillars implementation \cite{pcdetPP} utilizes cuDNN's dense convolution for the backbone layer. We compare the performance of our sparse convolution approach with that of the cuDNN dense convolution.
As shown in  Fig \ref{fig:performanceComparison}, our sparse convolution demonstrates significant performance improvements compared to the dense convolution in cuDNN. 
Notably, as the number of input channels increases, the performance gains become more significant. 
This is possibly because, when the number of input channels is small, the scattering step dominates the processing time, and its reliance on atomicAdd() operations results in slower performance. 
However, as the number of input channels grows, the matrix multiplication step becomes the dominant factor in processing time, benefiting from the efficient implementation provided by cuBLAS.

We implemented the backbone layer of the detection model using sparse convolution with a scatter operation and evaluated its performance on real LiDAR data consisting of 1,000 frames.
The backbone layer comprises three convolution blocks and three deconvolution (transposed convolution) operations. Each convolution block contains 4, 6, and 6 convolutions, respectively. Our sparse convolution is used to implement both the convolutions and deconvolutions within the backbone.
Experiments show that our sparse convolution reduces inference time by 28.5\%. However, this reduction falls short of our expectations. This is because, while the initial convolutions operate on sparse data, the majority of subsequent convolutions process dense data due to the expansion property inherent in the convolution operation. To mitigate the expansion caused by the convolution operation, the submanifold sparse convolution is employed in each convolution block, except for the first convolution of each block. 
The submanifold sparse convolution is implemented using the scatter operation.
Experiments show that inference can be accelerated by a factor of 2.3 when the submanifold sparse convolution is employed. 

\section{Model Performance Across Different Use Cases} \label{subsubsec:modelPerformance}

\subsection{Experiment 1:  Newcastle use case}\label{subsec:Newcastle}

\begin{figure}[ht]
\centering
\includegraphics[scale=0.29]{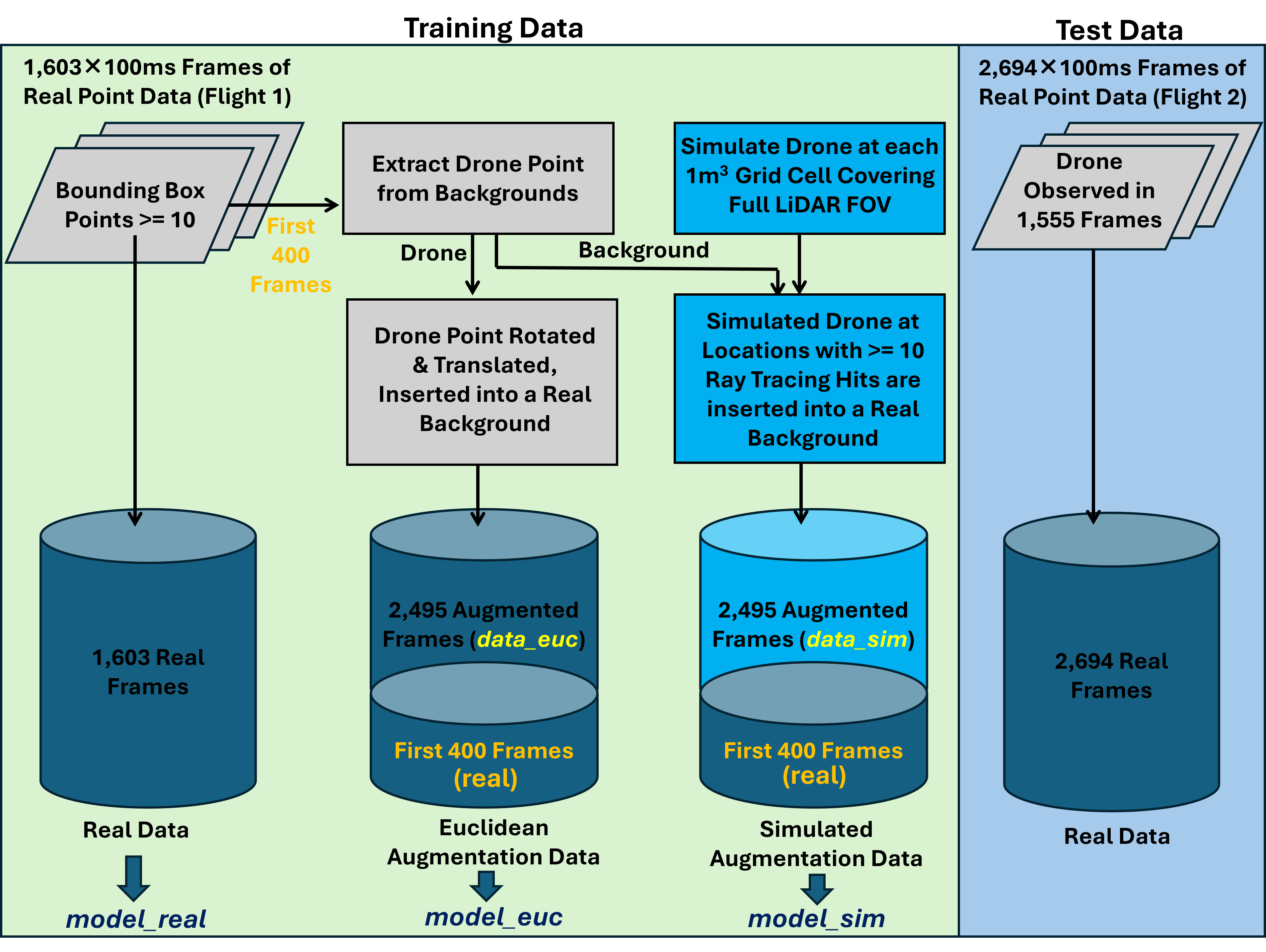}
\caption{Schematic diagram for dataset generation and model training.}
\label{fig:dataset-schematic}
\end{figure}

\begin{table*}
\centering
\caption{Evaluation of the detection model on Newcastle dataset}
\begin{tabular}{|c|c|c|c|c|c|}
  \hline
  \textbf{Trained model}  & \textbf{Training dataset}  & \textbf{Intensity value}  & \textbf{Precision}  & \textbf{Recall}  & \textbf{F1 score} \\
  \hline
   \multirow{4}{*}{model\_sim} &  \multirow{2}{*}{$(data\_sim)$ only}  & applied &0.511  &0.737  & 0.603\\
   \cline{3-6}
  &  & not applied &0.949  &0.788  & 0.861\\

 \cline{2-6}
 & \multirow{2}{*}{$(data\_sim)$ + $(real\_first\_400\_frame)$}  & applied &0.523  &0.75  &0.616 \\
   \cline{3-6}
  &  & not applied &0.96  &0.80  &0.872 \\
  \hline

 \multirow{2}{*}{ model\_euc} & \multirow{2}{*}{$(data\_euc)$ +  $(real\_first\_400\_frame)$}  & applied &0.24  &0.774 &0.367 \\
  \cline{3-6}
  &    & not applied &0.766  &0.513  &0.614 \\
  \hline
  model\_real &  1603 real frames collected from real drone fly   & applied & 0.826  & 0.62  & 0.706 \\
  \hline
\end{tabular}
\label{tab:newcastle-precision-recall}
\end{table*}

\begin{figure*}[ht]
    \centering
 \subfloat[\centering ]{{\includegraphics[width=3.65cm]{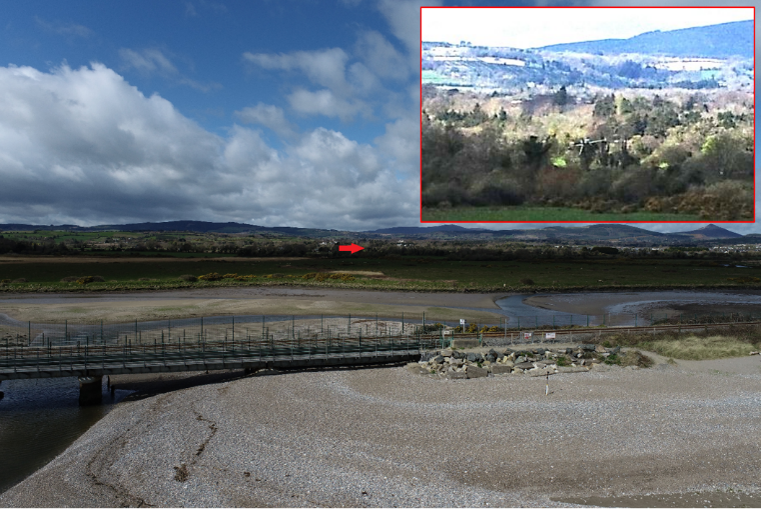} }}%
    \vspace{0.2cm}
    \qquad
 \subfloat[\centering ]{{\includegraphics[width=3.65cm]{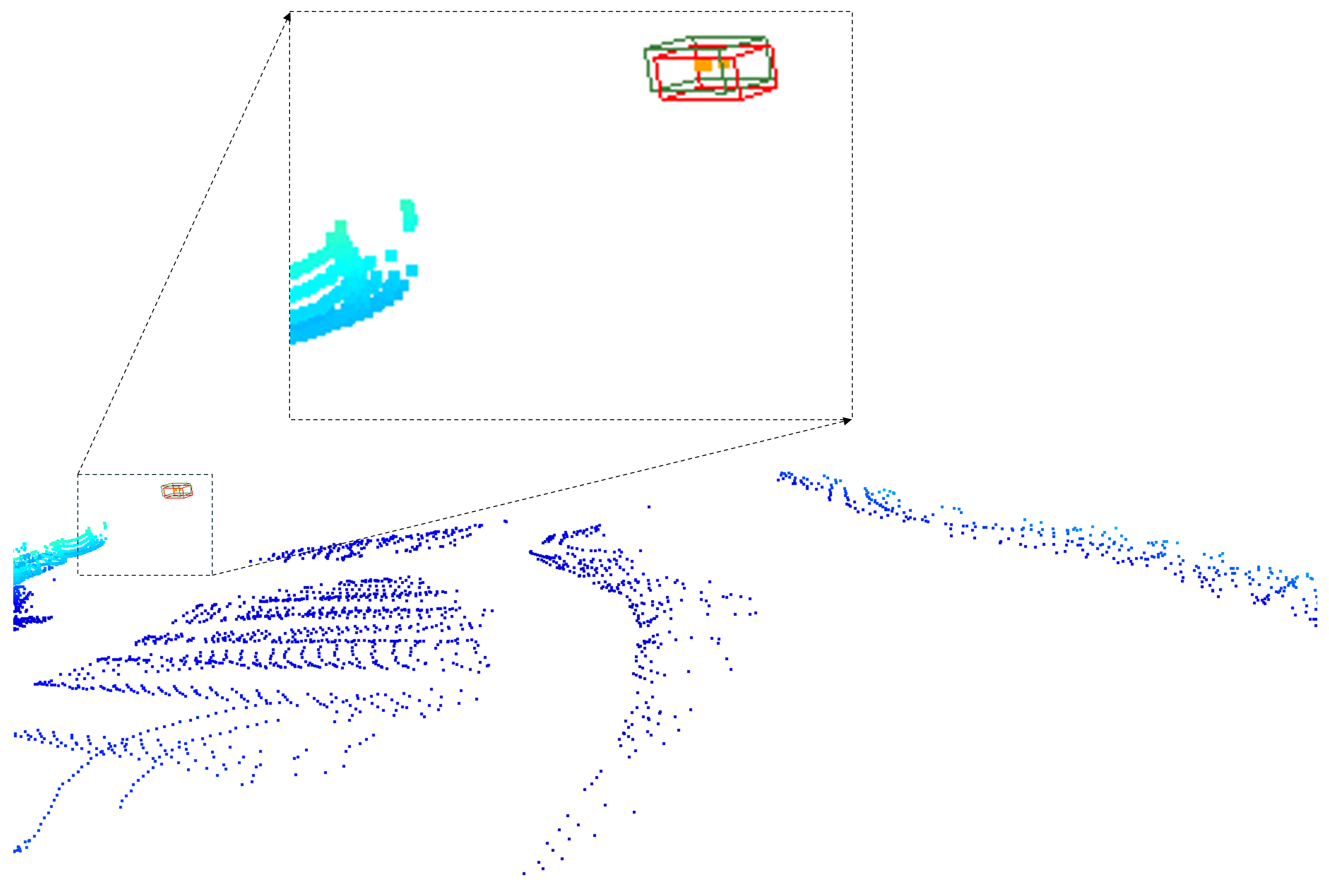} }}%
    \vspace{0.2cm}
    \qquad
 \subfloat[\centering ]{{\includegraphics[width=3.65cm]{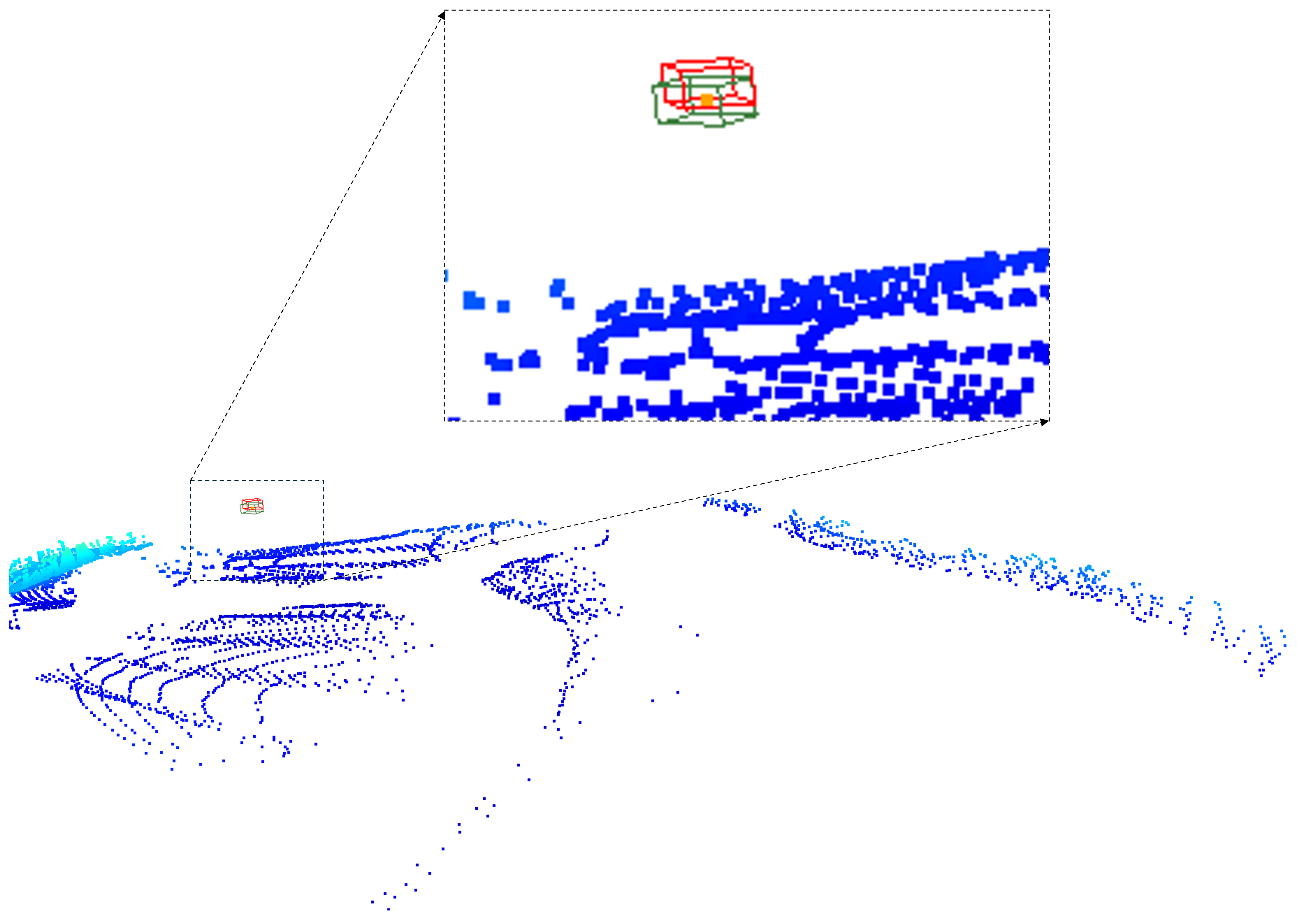} }}%
    \vspace{0.2cm}
    \qquad
 \subfloat[\centering ]{{\includegraphics[width=3.65cm]{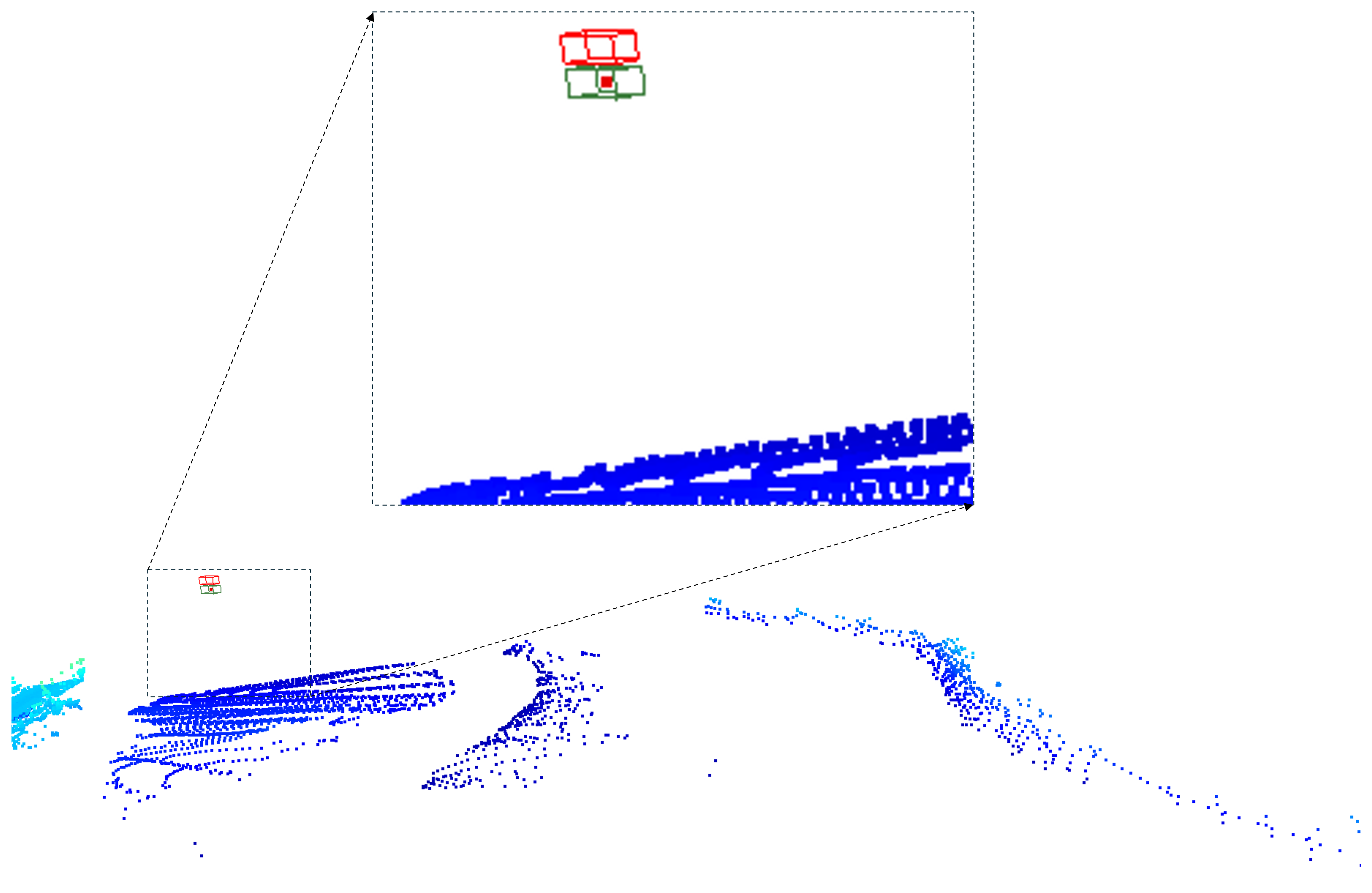} }}%
   \caption{Visualization of predicted bounding box (red) and ground truth bounding box (green), the inference model used is $(model\_sim)$ described in Section \ref{subsec:Newcastle}, a portion of the frame containing the drone is enlarged for clarity. (a)  Newcastle scenario: Two drones deployed to survey the railway bridge. The red arrow indicates the target drone in the image captured by the camera of the monitoring drone. The small picture inside the image is captured by the high resolution camera of photogrammetry drone. (b) The predicted bounding box has a good overlap with the ground truth bounding box (c) The predicted heading is different with the ground truth (d) The difference between the center of predicted bounding box and the center of ground truth bbox is significant.}
    \label{fig:newcastle-two-bbox}%
\end{figure*}

The objective in this experimentation with the Newcastle scenario was to evaluate the performance of the model and the data augmentation strategy. The data generation scheme and model training are depicted in Fig. \ref{fig:dataset-schematic}, with a detailed explanation provided in Appendix \ref{aug_scheme}. 
The training, validation, and testing split ratio is 2.4:1:2.2.

\subsubsection{Performance on real data} \label{sec:comparision}

We first train our 3D object detection model on real data without augmentation and compare its performance to detection models available in PCDet \cite{openpcdet2020}, an open-source toolbox designed for point cloud-based 3D object detection. PCDet is widely used for LiDAR-based perception tasks and supports multiple state-of-the-art detection methods. Its codebase is highly modular and includes both voxel-based and pillar-based models. For our comparison, we focus on pillar-based models from PCDet, as they do not impose strict restrictions on the Z-axis range, unlike voxel-based models. This flexibility allows us to adapt these models for drone-captured data with varying altitudes. Specifically, we compare our model with the state-of-the-art CenterPoint-PillarNet and DSVT-PillarNet models from PCDet, modifying their Z-axis range to better accommodate our drone data with varying altitudes. CenterPoint model reformulates detection as a center-based approach rather than relying on anchors or dense bounding box predictions. DSVT is an attention-based model that selectively processes only the important sparse voxel regions. Table \ref{tab:comp-other-model} presents a comparison between our model and the PCDet models, using a single NVIDIA RTX 3090 GPU. We used 1,603 real frames from the Newcastle dataset for training.

As shown in Table \ref{tab:comp-other-model}, CenterPoint-PillarNet achieves the highest precision but at the cost of slower inference speed compared to our model. DSVT-PillarNet performs poorly in both precision and inference time. Our model delivers the fastest inference speed. Specifically, the version of our model with both Z-axis range extension and (sparse+submanifold) convolution acceleration achieves the fastest inference time, with only a slight precision sacrifice (less than 1\%) compared to the model with Z-axis range extension alone. In our safety-critical application scenario, response time is crucial for real-time hazard avoidance. Therefore, we selected and modified PointPillars to prioritize faster inference times while maintaining acceptable precision.

\subsubsection{Inference time on embedded hardware}\label{sec:real-time-performance}

We evaluated the inference runtime on the Newcastle dataset. As shown in Table \ref{tab:inference-time}, for inference with FP32, the Pillarization part (Pillar Feature Net) accounts for approximately 1.8\% of the total inference time.
The Backbone + Head part is the most time-consuming, making up 81.8\% of the overall runtime. 
The Postprocess + NMS (Non-Maximum Suppression \cite{NMS}) step contributes around 16.4\% to the total inference time.

Regarding precision, the mixed precision implementation significantly reduced the overall runtime, particularly in the Backbone + Head. 
This shows that optimization through mixed precision making it a promising approach for real-time applications on embedded hardware like the Jetson Xavier.

\begin{table}[th]
\centering
\caption{Runtime of each model component on NVIDIA Jetson AGX Xavier.}
\begin{tabular}{|c|c|c|}
  \hline
 \textbf{Model Component} &  \makecell{ \textbf{Runtime (ms)}  \\ \textbf{(FP32)}} & \makecell{ \textbf{Runtime (ms)}\\ \textbf{(Mixed Precision)}} \\
  \hline
Pillarization & 1.21 & 0.82 \\
  \hline
Backbone + Head & 54.27 & 32.49 \\
  \hline
Postprocess + NMS & 10.89 &8.16 \\
  \hline
Overall & 66.37 & 41.47 \\
  \hline
\end{tabular}
\label{tab:inference-time}
\end{table}

\subsubsection{Performance on augmented data} \label{sec:aug}

As shown in Table \ref{tab:newcastle-precision-recall}, the model trained on a point cloud consisting of both synthetic augmentation data $(data\_sim)$ and real data from the first 400 frames $(real\_first\_400\_frame)$, without utilizing intensity values, achieves the best performance in all of the evaluation metrics, Precision, Recall, and F1 score. However, the model's performance decreases when intensity values are used for training, indicating that there is a discrepancy between the the simulated intensity values and the ground truth intensity values.
The model trained on a point cloud generated by combining Euclidean augmentation of real data, specifically $(data\_euc)$ and $(real\_first\_400\_frame)$, without utilizing intensity values, achieves a competitive performance.
The model's performance decreases when using intensity values for training, potentially due to the aforementioned possible discrepancy.
The model trained solely on simulated data $(data\_sim)$ shows a slight decrease in all metrics compared to the model trained on a combination of $(data\_sim)$ and $(real\_first\_400\_frame)$. This demonstrates that the model trained with simulated data alone can achieve performance remarkably close to that of the model trained with both simulated and real data.
The model trained on real data without any modifications achieves a comparable performance. However, it produces more false negatives, suggesting a lower capability for detecting drones.
These results demonstrate that our simulation augmentation strategy can generate more accurate data than Euclidean augmentation of real data.

Visualization of both the predicted bounding box and the ground truth bounding box is shown in Fig. \ref{fig:newcastle-two-bbox}, where the inference model applied is $(model\_sim)$. Through our testing, we discovered that predicting the heading of the detected drone is the most challenging task. This could be attributed to the fact that heading prediction typically requires a sufficient number of points representing the geometric structure of the body of the drone. However, in most cases, the LiDAR cannot generate an adequate number of points, because the drone is such a small object. 
In certain instances, as depicted in Fig. \ref{fig:newcastle-two-bbox}(d), there is a noticeable difference between the center of the predicted bounding box and that of the ground truth bounding box. This observation suggests that our model is imperfect.

\subsection{Experiment 2: Hamburg use case}\label{subsec:Hamburg}

\begin{figure}[ht]
\centering
\includegraphics[scale=0.44]{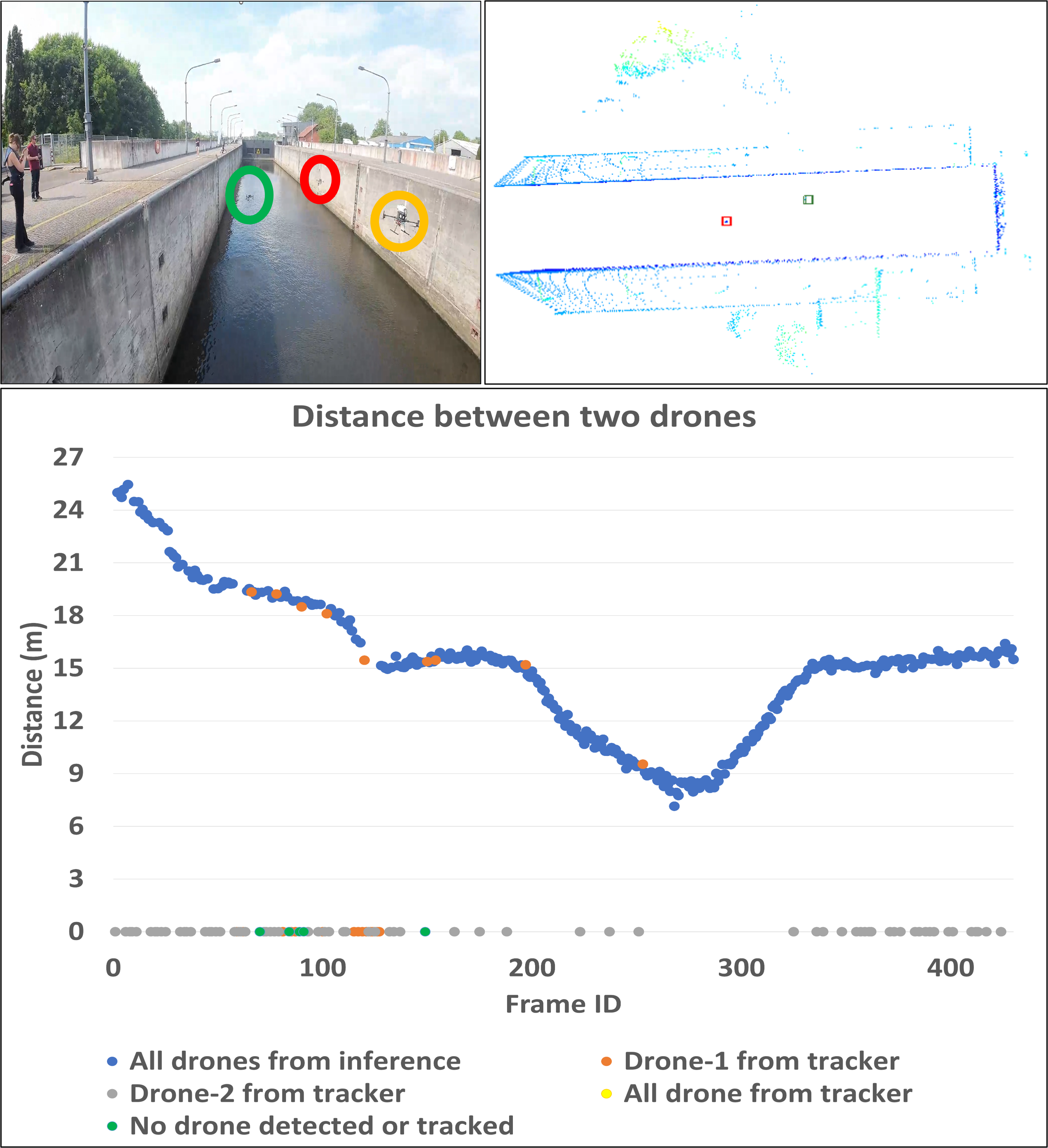}
\caption{Detection results at Harburg Lock. \textbf{Top left:} Harburg Lock scenario: Two drones (circled red and green) deployed for an inspection survey, the third drone is (circled yellow) monitoring their separation distance. \textbf{Top right:} Detection of two drones using the sense and detect system. The red bounding box is predicted by the deep-learning based detection model and the green bounding box is by the tracking algorithm. \textbf{Bottom:} Distance between the two drones fly-by, where the distance is computed using the detection results of detection and tracking algorithm.}
\label{fig:ham-two-bbox}
\end{figure}

This test involves an inspection survey with two drones (see the top left part of Fig. \ref{fig:ham-two-bbox}) simultaneously flying LiDAR (circled red) and camera (circled green) data capture missions.
The detection model was trained on data that was previously captured in the lock and augmented with synthetic observations using the digital twinning method described earlier. The 100ms point cloud scenes presented to the detection model embedded on the monitoring drone typically contained observations of the LiDAR drone positioned relatively high and well clear of other objects, as well as the camera drone positioned relatively low and near the concrete wall. 

The sense and detect system combined the deep learning based detection model with the tracking-by-detection algorithm described earlier. The lower part of Fig. \ref{fig:ham-two-bbox} displays a graph of the distance between the two drones during an example close proximity fly-by.
The quantitative data is presented in Table \ref{tab:distance-graph}. There are 430 frames in total, representing 43 seconds of flight perception data and translating to an average relative speed of c. 1m/s between the two drones.

As shown, the combination of the deep-learning based detection model and the detection-by-tracking algorithm enables the sense and detect system to robustly track the positions of two drones. The solution accuracy is the range resolution of the LiDAR, in this case 5cm, which is comparable with RTK GPS solutions. As it is a relative based positioning and deconfliction system it is robust to drop out of satellite coverage, while the ability to automatically track dynamic objects compares favorably to a ground based total station.

\begin{table}[th]
\centering
\caption{Quantitative data associated with graph in Fig. \ref{fig:ham-two-bbox}.}
\begin{tabular}{|c|c|}
  \hline
  Two drones tracked by inference model detection &  333 \\
  \hline
  \makecell{Two drones tracked by combination of inference model (1) \\ and tracker algorithm (1)} & 9  \\
  \hline
  Only one drone tracked by either method &   83\\
  \hline
  No drones detected or tracked &  5 \\
  \hline
\end{tabular}
\label{tab:distance-graph}
\end{table}


%% file: sec/discusstion.tex
\section{Discussion}

\subsection {Detection Performance and Real-World Feasibility}

This work presents the first airborne LiDAR-based deep learning detection framework designed to operate entirely onboard an airborne embedded platform, enabling real-time detection without reliance on ground-based processing. The results demonstrate that our model achieves high recall and precision in airborne conditions, even in complex operational environments typical of transport system infrastructure.
Our findings show that incorporating Digital Twin-based augmentation significantly improves detection robustness, with models trained on augmented data achieving a recall improvement of 18\% and precision improvement of 13\% compared to conventional augmentation techniques. The ability to enhance model generalization through physics-informed augmentation is crucial for real-world deployment, where training data diversity is often limited.

\subsection {Sparse Convolution Acceleration and Embedded Efficiency}

A major challenge in deploying deep learning-based airborne detection is the severe computational constraints imposed by embedded systems. Unlike automotive applications, where dedicated high-power GPUs are available, airborne platforms must balance detection accuracy with real-time inference speed while operating under strict size, weight, and power (SWaP) constraints.
Our scatter-based sparse convolution approach addresses this challenge, reducing inference latency by 2.3× while maintaining competitive accuracy. The observed efficiency gains scale with increasing input channels, as the optimized implementation shifts the computational burden toward matrix multiplication, benefiting from cuBLAS acceleration. This result confirms that real-time LiDAR-based perception is feasible onboard airborne systems without requiring specialized hardware accelerators.

\subsection {Challenges in Deploying LiDAR Detection in Constrained Environments}

The field deployments highlighted several practical challenges for airborne LiDAR-based detection in real-world infrastructure monitoring. The detection performance was affected by targets frequently appearing at the periphery of the LiDAR scan region, reducing effective point density. Structural elements introduced multipath and erroneous returns, requiring additional filtering and tracking techniques. Unlike static infrastructure, the presence of moving vessels, water reflections, and confined spaces challenged the detection consistency. These results indicate that while airborne LiDAR detection is feasible, some adaptations are required for deployment in enclosed or highly dynamic operational settings.

%% file: sec/conclusion.tex
\section{Conclusion}
This paper tackles the challenge of monitoring a drone swarm during collaborative tasks.
It presents
the first
airborne LiDAR based solution for drone-swarm detection and localization
using a 3D deep learning model.
It adapts an existing detection model, originally designed for ground applications, to an air-to-air drone scenario by expanding the scan space vertically. 
To reduce the latency of the detection model, we propose a new sparse convolution algorithm that improves efficiency by bypassing the mapping between input and output, which is required by existing methods.
This algorithm accelerates the model's backbone layer, which is the most time-consuming component of the detection process.
A scenario Digital Twin is implemented to simulate close encounters and safety-critical scenarios to gather training data. By augmenting the actual background data with high-quality synthetic drone data, we improve the accuracy and efficiency of our training and inferencing processes.
Moreover, we incorporate a tracking-by-detection algorithm into the system, which allows for accurate monitoring of the separation distance of multiple drones, even in challenging environments.

Our approach has been successfully validated through real-world tests, with the trained model achieving an impressive 80{\%} recall and 96{\%} precision when tested on real datasets. We trained the detection model on several datasets, each generated using different augmentation strategies, and found that our synthetic augmentation strategy produced more precise data than the conventional augmentation method used with real data.
Overall, our proposed approach offers a novel and effective solution for drone swarm detection and localization, with promising results from our tests.

%

%% file: sec/bio.tex
\bibliographystyle{IEEEtran} 
\bibliography{relatedworks} 

\setlength{\parskip}{1em}

%% file: sec/appendix.tex
\section{LiDAR simulation with NVIDIA Optix Ray Tracing} \label{appendix:sim}
We incorporate the NVIDIA OptiX Ray Tracing engine \cite{Optix} to accelerate the simulation of the LiDAR, using a GPU system equipped with Ray Tracing cores. The OptiX engine utilizes hardware-accelerated data structures, such as the Bounding Volume Hierarchy (BVH), to expedite the ray tracing process. This technology enables the OptiX engine to efficiently traverse complex scenes and accurately compute the intersection of rays with objects.
During the simulation process, the mesh model of the target drone will be repeatedly rotated and translated within the LiDAR's field of view, requiring frequent rebuilding or updating of the acceleration data structure. However, this rebuilding or updating process can be computationally expensive.
We propose an alternative approach that eliminates the need for rebuilding or updating the acceleration data structure. This method allows for a significant reduction in the computational cost of the simulation process.

In the conventional approach, the drone is first rotated, and then the LiDAR ray is projected onto it. This requires rebuilding or updating the acceleration structure within the ray tracing core.
However, since rotation and translation are relative in physics, we can instead rotate and translate the LiDAR ray's origin and direction vector, avoiding the need to rotate and translate the drone mesh model.
Specifically, we first align the origin of the LiDAR ray with the origin of the 3D coordinate system, and then align the centerline of the LiDAR's field of view with the x-axis. Assuming the rotation angle of the drone is $\theta$ (note that we do not actually rotate the drone mesh in our implementation), the proposed LiDAR simulation, which bypasses the need to rebuild or update the acceleration structure, involves the following steps:
\begin{itemize}
\item \textbf{Step 1.} Rotate the direction vectors of the LiDAR ray -$\theta$ degrees around z-axis. 
\item \textbf{Step 2.} Move the origin of the coordinate system to the center of the drone mesh model.
\item \textbf{Step 3.} Rotate the origin of the LiDAR ray -$\theta$ degrees around z-axis. 
\item \textbf{Step 4.} Move the origin of the coordinate system to the origin of LiDAR ray. 
\item \textbf{Step 5.} Compute the intersection points between LiDAR rays and drone triangle mesh using OptiX.
\end{itemize}
Each OptiX thread computes a new ray origin and direction vector independently, rather than rebuilding the acceleration structure in the ray tracing core.


For the 3D rotation of a drone triangle mesh, we only consider rotation around the $z$-axis (up-right). The following rotation matrix is used, where $\theta$ is the angle of rotation.
$$
\begin{bmatrix}
\cos(\theta) & -\sin(\theta) & 0 \\
\sin(\theta) & \cos(\theta) & 0 \\
0 & 0 & 1
\end{bmatrix}
$$
\quad

The reflection model used by the DJI L1 Lidar is based on the Lambertian reflection model \cite{losasso2004surface}. This model assumes that the intensity of the reflected laser beam is proportional to the cosine of the angle between the laser beam and the surface normal of the object being scanned, 
\begin{equation}
I = IO \times cos(\alpha)
\end{equation}
where $I$ is the intensity of the reflected ray, $IO$ is the intensity of the incident ray, and $\alpha$ is the angle between the incident ray and the surface normal.

\section{Evaluation Metrics}  \label{appendix_metric}

Recall ($R$) is the ratio of the number of true positive ($TP$) examples to the sum of true positives and false negatives ($FN$) examples. It measures the model's ability to correctly identify all positive examples out of all the positive examples in the dataset.

\textcolor{black}{
\begin{equation}
R = \frac{TP}{TP + FN}
\end{equation}
}

Precision ($P$) is the ratio of the number of true positive ($TP$) examples to the sum of true positives and false positives ($FP$) examples. It measures the model's ability to identify only the relevant positive examples out of all the positive predictions. 

\textcolor{black}{
\begin{equation}
P = \frac{TP} {TP + FP}
\end{equation}
}

The F1 score is the harmonic mean of precision and recall. It is a single metric that balances both precision and recall. A high F1 score means that the model has both high precision and high recall.

\textcolor{black}{
\begin{equation}
F1 = 2 \times  \frac {P \times R}{P + R}
\end{equation}
}

The recall, precision, and F1 score are calculated using IoU (Intersection over Union) scores, which is a measure of the overlap between the predicted bounding box and the ground truth bounding box. The intersection area is the area where the predicted and ground truth regions overlap, while the union area is the total area of both regions. The IoU score is then calculated as:

\textcolor{black}{
\begin{equation}
IoU = \frac{Intersection}{Union}
\end{equation}
}

\section {Data generation scheme} \label {aug_scheme}

We collect training data by scanning the target drone (detection object) using a LiDAR sensor mounted on a monitoring drone. This process begins with the target drone positioned 10 meters away from the monitoring drone, after which it flies along the x-axis.
It is important to note that the LiDAR sensor serves as the origin of the 3D coordinate system and is oriented along the x-axis. As demonstrated in the paper, the non-uniform scan pattern of the LiDAR sensor results in the highest point density when the target drone moves along the x-axis.
Once the target drone reaches a given point, it will change elevation.

The collected LAS dataset $(data\_orig)$ 
was segmented into 100ms frames, and the target drone observations were labelled with bounding boxes placed at locations derived from the target drone flight log. The dataset was then reduced into the set of frames (N=1603) that contained bounding boxes enclosing at least 10 points. The first 400 instances $(real\_first\_400\_frame)$ of the resulting frames were filtered to extract observations of the target drone and the background separately. We call them $(extracted\_drone)$ and $(extracted\_background)$ respectively. During data collection, these 400 frames were recorded while the target drone moves along the x-axis. 
This approach ensures that each frame contains the densest point clouds, effectively preserving the geometric shape of the drone from various positions. On the other hand, the first 400 frames are recorded right after IMU calibration. Frames recorded right after IMU calibration will have the tightest positional accuracy for labelling. Frames could be chosen from later in the dataset after later IMU calibrations, but the first 400 are the easiest timestamp wise to utilise. By augmenting these 400 frames, we can generate a high-quality, geometry-preserving augmented dataset, improving the robustness and reliability of the training data.

Our augmentation strategy simulates LiDAR scans by placing a 3D drone mesh at randomized orientations within a stratified FOV grid and performing ray tracing over 100ms to generate physically accurate point clusters, retaining only those with at least ten valid hits.
Each of the 2495 instances was then randomly matched to one of the 400 backgrounds $(extracted\_background)$. The resulting dataset $(data\_sim)$ and the real 400 frames $(real\_first\_400\_frame)$ were used together to train the detection model $(model\_sim)$. The set of insertions locations was retained and reused in generating a comparative dataset from Euclidean augmentations in the second strategy.

To generate the Euclidean augmentation dataset $(data\_euc)$, each instance of $(extracted\_drone)$ were upsampled into 2895 exemplars by creating 2495 additional rotated copies. Each point cluster was then translated to the corresponding background and insertion locations retained during construction of the $data\_sim$ dataset. The resulting dataset $(data\_euc)$ was used to train the detection model $(model\_euc)$.

Both training datasets are the same size, contain the same number of augmentations, and corresponding frames have the same background with a drone inserted in the same location.
To test the performance of the respective models, we collected new LiDAR data from the same mission site, which consists of consecutive 2694 frames. In these 2694 frames, drones (ground truth) observed in 1555 frames.

\section{Survey sites}\label{survey-sites}
The data specification of maintenance inspection surveys facilitates swarm flights as multiple drones can simultaneously fly spatially segmented waypoint- and route-based data collection missions as the different point- and pixel- datasets have different resolution requirements. For example, collection of sub millimeter Ground Sample Density camera imagery in photogrammetry and crack detection use cases can require the drone to fly within 2-6m of the target structure etc., while LiDAR surveys aiming for centimeter point density can fly at offset distances of 20-30m. This facilitates separation and deconfliction of planned swarm trajectories during strategic flight risk management. 

\subsection{Newcastle scenario}\label{app:newcastle}
\begin{figure}[ht]
\centering
\includegraphics[scale=0.30]{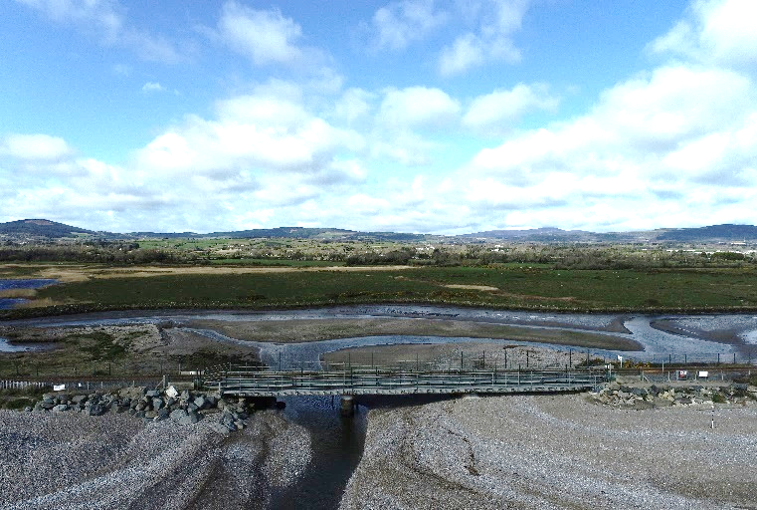}
\caption{Test site 1: The railway bridge 1.5km north of Newcastle Aerodrome (EINC), County Wicklow, Ireland.}
\label{fig:newcastle}
\end{figure}

The Newcastle scenario see Fig. \ref{fig:newcastle} involved two drones deployed to survey the railway bridge, one drone carrying a LiDAR sensor and the other with a high-resolution camera for photogrammetry. The LiDAR flight was approximately 10 minutes in duration, including IMU calibrations at the start and end of the mission. The LiDAR drone frequently observed the camera drone, with observations in the point cloud data annotated using positional information extracted from the flight logs.

\subsection{Hamburg scenario}\label{app:hamburg}

\begin{figure}[!t]
\centering
\includegraphics[scale=0.26]{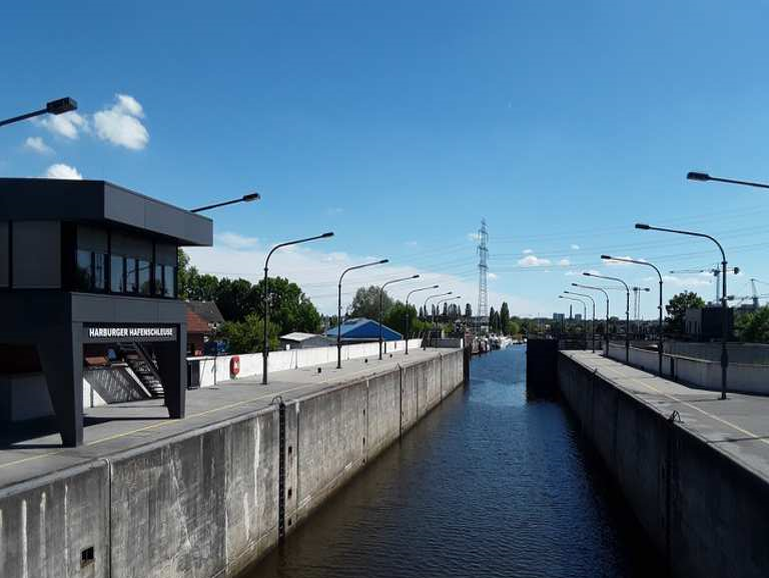}
\caption{Test site 2: The Harburger-Lock, Hamburg, Germany.}
\label{fig:hamburg}
\end{figure}

Testing at Harburg Lock, Hamburg, Germany (see Fig. \ref{fig:hamburg}) involved an inspection survey with two drones simultaneously flying LiDAR and camera data capture missions. The camera drone
mission
was 
to capture imagery of the lock wall with c. 1mm Ground Sample Distance, flying 
within the lock chamber to a distance of 1-2m of the lock wall (equipped with a 45MP camera) while oriented towards the wall. 
The LiDAR drone flew at a horizontal distance of approximately 10m from the camera drone, following a systematic back-and-forth flight pattern.

The surrounding environment was complex, cluttered, and dynamic encompassing people, buildings, trees and vegetation, as well as a wide range of building and man-made objects (e.g. lamp posts). Both drones tended to appear at the periphery of the usable field of view of the LiDAR of the monitoring drone.

While the flight plans for both drones were strategically separated using trajectory-based planning, the camera drone was required to fly a runline that risked denial of GNSS and communications due to multi-path and signal occlusion by the walls of the lock chamber. The operational risk management incorporated additional controls to monitor and track the drones during scheduled close encounters and to mitigate against potential air-to-air collision in the event of unplanned deviations. Potential trigger events included incursions by people or vehicles and return-to-home events due to battery depletion. 

A drone equipped with a Livox Avia LiDAR and the sense and detect system was deployed to monitor the two survey drones. The system was configured to raise an alert once a minimum separation threshold was breached, in this instance when the survey drones came within 15m of each other. The monitoring drone was positioned to station keep at one end of
the Lock and at a flight elevation midway between the two survey drones.


%% file: revised_version_1.bbl
\begin{thebibliography}{10}
\providecommand{\url}[1]{#1}
\csname url@samestyle\endcsname
\providecommand{\newblock}{\relax}
\providecommand{\bibinfo}[2]{#2}
\providecommand{\BIBentrySTDinterwordspacing}{\spaceskip=0pt\relax}
\providecommand{\BIBentryALTinterwordstretchfactor}{4}
\providecommand{\BIBentryALTinterwordspacing}{\spaceskip=\fontdimen2\font plus
\BIBentryALTinterwordstretchfactor\fontdimen3\font minus
  \fontdimen4\font\relax}
\providecommand{\BIBforeignlanguage}[2]{{%
\expandafter\ifx\csname l@#1\endcsname\relax
\typeout{** WARNING: IEEEtran.bst: No hyphenation pattern has been}%
\typeout{** loaded for the language `#1'. Using the pattern for}%
\typeout{** the default language instead.}%
\else
\language=\csname l@#1\endcsname
\fi
#2}}
\providecommand{\BIBdecl}{\relax}
\BIBdecl

\bibitem{IEEEsenseavoid}
G.~Fasano, D.~Accado, A.~Moccia, and D.~Moroney, ``{Sense and avoid for
  unmanned aircraft systems},'' \emph{IEEE Aerospace and Electronic Systems
  Magazine}, vol.~31, no.~11, pp. 82--110, 2016.

\bibitem{Gkoumas02082024}
I.~C. Konstantinos~Gkoumas, Marcin~Stepniak and F.~M. dos Santos, ``New
  technologies for bridge inspection and monitoring: a perspective from
  european union research and innovation projects,'' \emph{Structure and
  Infrastructure Engineering}, vol.~20, no. 7-8, pp. 1120--1132, 2024.

\bibitem{10399955}
M.~C. Santos, B.~Bartlett, V.~E. Schneider, F.~O. Bradaigh, B.~Blanck, P.~C.
  Santos, P.~Trslic, J.~Riordan, and G.~Dooly, ``Cooperative unmanned aerial
  and surface vehicles for extended coverage in maritime environments,''
  \emph{IEEE Access}, vol.~12, pp. 9206--9219, 2024.

\bibitem{thales}
N.~Vigne, R.~Barrère, B.~Blanck, F.~Steffens, C.~N. Au, J.~Riordan, and
  G.~Dooly, ``Embedded port infrastructure inspection using artificial
  intelligence,'' in \emph{OCEANS 2023 - Limerick}, 2023, pp. 1--8.

\bibitem{adam2024risks}
J.~M. Adam, S.~Makoond, F.~Xu, and H.~Ye, ``Risks of bridge collapses are real
  and set to rise — here's why,'' \emph{Nature}, vol. 629, pp. 1001--1003,
  2024.

\bibitem{ali2019artificial}
G.~Ali, A.~Elsayegh, R.~Assaad, I.~H. El-Adaway, and I.~S. Abotaleb,
  ``Artificial neural network model for bridge deterioration and assessment,''
  2019.

\bibitem{GPSdrift}
K.~Kim, W.~Kim, D.~Choi, and H.~Myung, ``{Calibration of the drift error in GPS
  using optical flow and fixed reference station},'' in \emph{2015 15th
  International Conference on Control, Automation and Systems (ICCAS)}, 2015,
  pp. 1370--1373.

\bibitem{hommes2016detection}
A.~Hommes, A.~Shoykhetbrod, D.~Noetel, S.~Stanko, M.~Laurenzis, S.~Hengy, and
  F.~Christnacher, ``Detection of acoustic, electro-optical and radar
  signatures of small unmanned aerial vehicles,'' in \emph{Target and
  Background Signatures II}, vol. 9997.\hskip 1em plus 0.5em minus 0.4em\relax
  SPIE, 2016, p. 999701.

\bibitem{acoustic1}
D.~Tejera-Berengue, F.~Zhu-Zhou, M.~Utrilla-Manso, R.~Gil-Pita, and
  M.~Rosa-Zurera, ``{Acoustic-Based Detection of UAVs Using Machine Learning:
  Analysis of Distance and Environmental Effects},'' in \emph{2023 IEEE Sensors
  Applications Symposium (SAS)}.\hskip 1em plus 0.5em minus 0.4em\relax IEEE,
  2023, pp. 1--6.

\bibitem{RF1}
P.~Nguyen, V.~Kakaraparthi, N.~Bui, N.~Umamahesh, N.~Pham, H.~Truong,
  Y.~Guddeti, D.~Bharadia, R.~Han, E.~Frew, D.~Massey, and T.~Vu,
  ``{DroneScale: drone load estimation via remote passive RF sensing},'' in
  \emph{Proceedings of the 18th Conference on Embedded Networked Sensor
  Systems}.\hskip 1em plus 0.5em minus 0.4em\relax New York, NY, USA:
  Association for Computing Machinery, 2020, p. 326–339.

\bibitem{radar1}
M.~Caris, S.~Stanko, W.~Johannes, S.~Sieger, and N.~Pohl, ``Detection and
  tracking of micro aerial vehicles with millimeter wave radar,'' in \emph{2016
  European Radar Conference (EuRAD)}, 2016, pp. 406--408.

\bibitem{radar2}
{\'A}.~D. de~Quevedo, F.~I. Urzaiz, J.~G. Menoyo, and A.~A. L{\'o}pez, ``Drone
  detection and radar-cross-section measurements by rad-dar,'' \emph{IET Radar,
  Sonar \& Navigation}, vol.~13, no.~9, pp. 1437--1447, 2019.

\bibitem{radarHector}
H.~Arroyo, P.~Keir, D.~Angus, S.~Matalonga, S.~Georgiev, M.~Goli, G.~Dooly, and
  J.~Riordan, ``Segmentation of drone collision hazards in airborne radar point
  clouds using pointnet,'' \emph{IEEE Transactions on Intelligent
  Transportation Systems}, vol.~25, no.~11, pp. 17\,762--17\,777, 2024.

\bibitem{aker2017using}
C.~Aker and S.~Kalkan, ``Using deep networks for drone detection,'' in
  \emph{2017 14th IEEE International Conference on Advanced Video and Signal
  Based Surveillance (AVSS)}.\hskip 1em plus 0.5em minus 0.4em\relax IEEE,
  2017, pp. 1--6.

\bibitem{Detec-small-UVAs-pami}
A.~Rozantsev, V.~Lepetit, and P.~Fua, ``Detecting flying objects using a single
  moving camera,'' \emph{IEEE Transactions on Pattern Analysis and Machine
  Intelligence}, vol.~39, no.~5, pp. 879--892, 2017.

\bibitem{unlu2019deep}
E.~Unlu, E.~Zenou, N.~Riviere, and P.-E. Dupouy, ``Deep learning-based
  strategies for the detection and tracking of drones using several cameras,''
  \emph{IPSJ Transactions on Computer Vision and Applications}, vol.~11, no.~1,
  pp. 1--13, 2019.

\bibitem{detectFromVideo}
J.~Li, D.~H. Ye, M.~Kolsch, J.~P. Wachs, and C.~A. Bouman, ``Fast and robust
  uav to uav detection and tracking from video,'' \emph{IEEE Transactions on
  Emerging Topics in Computing}, vol.~10, no.~3, pp. 1519--1531, 2022.

\bibitem{StereoVisionDepth}
J.~Z. Sasiadek and M.~J. Walker, ``Achievable stereo vision depth accuracy with
  changing camera baseline,'' in \emph{2019 24th International Conference on
  Methods and Models in Automation and Robotics (MMAR)}, 2019, pp. 152--157.

\bibitem{Hammer2020AMA}
M.~Hammer, B.~Borgmann, M.~Hebel, and M.~Arens, ``A multi-sensorial approach
  for the protection of operational vehicles by detection and classification of
  small flying objects,'' in \emph{Security + Defence}, 2020.

\bibitem{causa2022closed}
F.~Causa, R.~Opromolla, and G.~Fasano, ``{Closed loop integration of air-to-air
  visual measurements for cooperative UAV navigation in GNSS challenging
  environments},'' \emph{Aerospace Science and Technology}, vol. 130, p.
  107947, 2022.

\bibitem{PC-Survey-2021}
Y.~Guo, H.~Wang, Q.~Hu, H.~Liu, L.~Liu, and M.~Bennamoun, ``{Deep Learning for
  3D Point Clouds: A Survey},'' \emph{IEEE Transactions on Pattern Analysis and
  Machine Intelligence}, vol.~43, no.~12, pp. 4338--4364, 2021.

\bibitem{PointNet}
R.~Charles, H.~Su, M.~Kaichun, and L.~J. Guibas, ``{PointNet: Deep Learning on
  Point Sets for 3D Classification and Segmentation},'' in \emph{2017 IEEE
  Conference on Computer Vision and Pattern Recognition (CVPR)}, Los Alamitos,
  CA, USA, 2017, pp. 77--85.

\bibitem{PointNet++}
C.~R. Qi, L.~Yi, H.~Su, and L.~J. Guibas, ``{PointNet++: Deep Hierarchical
  Feature Learning on Point Sets in a Metric Space},'' in \emph{Proceedings of
  the 31st International Conference on Neural Information Processing Systems},
  2017, pp. 5105--5114.

\bibitem{Voxelnet}
Y.~Zhou and O.~Tuzel, ``{VoxelNet: End-to-end learning for point cloud based 3d
  object detection},'' in \emph{Proceedings of the IEEE conference on computer
  vision and pattern recognition}, 2018, pp. 4490--4499.

\bibitem{second}
Y.~Yan, Y.~Mao, and B.~Li, ``{SECOND: Sparsely Embedded Convolutional
  Detection},'' \emph{Sensors}, vol.~18, no.~10, 2018.

\bibitem{deng2021voxel}
J.~Deng, S.~Shi, P.~Li, W.~Zhou, Y.~Zhang, and H.~Li, ``{Voxel r-cnn: Towards
  high performance voxel-based 3d object detection},'' in \emph{Proceedings of
  the AAAI Conference on Artificial Intelligence}, vol.~35, no.~2, 2021, pp.
  1201--1209.

\bibitem{PointPillars}
A.~H. Lang, S.~Vora, H.~Caesar, L.~Zhou, J.~Yang, and O.~Beijbom,
  ``{Pointpillars: Fast encoders for object detection from point clouds},'' in
  \emph{Proceedings of the IEEE/CVF Conference on Computer Vision and Pattern
  Recognition (CVPR)}, 2019, pp. 12\,697--12\,705.

\bibitem{SSD}
W.~Liu, D.~Anguelov, D.~Erhan, C.~Szegedy, S.~Reed, C.-Y. Fu, and A.~C. Berg,
  ``{SSD: Single shot multibox detector},'' in \emph{Computer Vision--ECCV
  2016: 14th European Conference, Amsterdam, The Netherlands, October 11--14,
  2016, Proceedings, Part I 14}.\hskip 1em plus 0.5em minus 0.4em\relax
  Springer, 2016, pp. 21--37.

\bibitem{shi2020pvrcnn}
S.~Shi, X.~Wang, and H.~Li, ``{PV-RCNN: Point-Voxel Feature Set Abstraction for
  3D Object Detection},'' in \emph{Proceedings of the IEEE/CVF Conference on
  Computer Vision and Pattern Recognition}, 2020, pp. 10\,529--10\,538.

\bibitem{yin2021center}
T.~Yin, X.~Zhou, and P.~Krahenbuhl, ``Center-based 3d object detection and
  tracking,'' in \emph{Proceedings of the IEEE/CVF conference on computer
  vision and pattern recognition}, 2021, pp. 11\,784--11\,793.

\bibitem{DSVT}
H.~Wang, C.~Shi, S.~Shi, M.~Lei, S.~Wang, D.~He, B.~Schiele, and L.~Wang,
  ``Dsvt: Dynamic sparse voxel transformer with rotated sets,'' in
  \emph{Proceedings of the IEEE/CVF Conference on Computer Vision and Pattern
  Recognition (CVPR)}, June 2023, pp. 13\,520--13\,529.

\bibitem{backbonetypeselection}
K.~Lis and T.~Kryjak, ``Pointpillars backbone type selection for fast and
  accurate lidar object detection,'' 2022.

\bibitem{graham2014spatially}
B.~Graham, ``Spatially-sparse convolutional neural networks,'' \emph{arXiv
  preprint arXiv:1409.6070}, 2014.

\bibitem{spcnn}
B.~Liu, M.~Wang, H.~Foroosh, M.~Tappen, and M.~Penksy, ``Sparse convolutional
  neural networks,'' in \emph{2015 IEEE Conference on Computer Vision and
  Pattern Recognition (CVPR)}, 2015, pp. 806--814.

\bibitem{Submanifold}
B.~Graham, M.~Engelcke, and L.~van~der Maaten, ``3d semantic segmentation with
  submanifold sparse convolutional networks,'' 2017.

\bibitem{cudnn-arxiv}
S.~Chetlur, C.~Woolley, P.~Vandermersch, J.~Cohen, J.~Tran, B.~Catanzaro, and
  E.~Shelhamer, ``cudnn: Efficient primitives for deep learning,'' \emph{arXiv
  preprint arXiv:1410.0759}, 2014.

\bibitem{minkowski}
C.~Choy, J.~Gwak, and S.~Savarese, ``4d spatio-temporal convnets: Minkowski
  convolutional neural networks,'' 2019.

\bibitem{torchsparse}
H.~Tang, S.~Yang, Z.~Liu, K.~Hong, Z.~Yu, X.~Li, G.~Dai, Y.~Wang, and S.~Han,
  ``Torchsparse++: Efficient training and inference framework for sparse
  convolution on gpus,'' in \emph{IEEE/ACM International Symposium on
  Microarchitecture (MICRO)}, 2023.

\bibitem{deltacnn}
M.~Parger, C.~Tang, C.~D. Twigg, C.~Keskin, R.~Wang, and M.~Steinberger,
  ``Deltacnn: End-to-end cnn inference of sparse frame differences in videos,''
  in \emph{2022 IEEE/CVF Conference on Computer Vision and Pattern Recognition
  (CVPR)}, 2022, pp. 12\,487--12\,496.

\bibitem{KITTI}
A.~Geiger, P.~Lenz, C.~Stiller, and R.~Urtasun, ``{Vision meets robotics: The
  kitti dataset},'' \emph{The International Journal of Robotics Research},
  vol.~32, no.~11, pp. 1231--1237, 2013.

\bibitem{Waymo}
P.~Sun, H.~Kretzschmar, X.~Dotiwalla, A.~Chouard, V.~Patnaik, P.~Tsui, J.~Guo,
  Y.~Zhou, Y.~Chai, B.~Caine \emph{et~al.}, ``Scalability in perception for
  autonomous driving: Waymo open dataset,'' in \emph{Proceedings of the
  IEEE/CVF conference on computer vision and pattern recognition}, 2020, pp.
  2446--2454.

\bibitem{nuScenes}
H.~Caesar, V.~Bankiti, A.~H. Lang, S.~Vora, V.~E. Liong, Q.~Xu, A.~Krishnan,
  Y.~Pan, G.~Baldan, and O.~Beijbom, ``{nuscenes: A multimodal dataset for
  autonomous driving},'' in \emph{Proceedings of the IEEE/CVF conference on
  computer vision and pattern recognition}, 2020, pp. 11\,621--11\,631.

\bibitem{abu2018augmented}
H.~Abu~Alhaija, S.~K. Mustikovela, L.~Mescheder, A.~Geiger, and C.~Rother,
  ``{Augmented reality meets computer vision: Efficient data generation for
  urban driving scenes},'' \emph{International Journal of Computer Vision},
  vol. 126, pp. 961--972, 2018.

\bibitem{ros2016synthia}
{Ros, German and Sellart, Laura and Materzynska, Joanna and Vazquez, David and
  Lopez, Antonio M}, ``The synthia dataset: A large collection of synthetic
  images for semantic segmentation of urban scenes,'' in \emph{Proceedings of
  the IEEE conference on computer vision and pattern recognition}, 2016, pp.
  3234--3243.

\bibitem{richter2016playing}
S.~R. Richter, V.~Vineet, S.~Roth, and V.~Koltun, ``Playing for data: Ground
  truth from computer games,'' in \emph{Computer Vision--ECCV 2016: 14th
  European Conference, Amsterdam, The Netherlands, October 11-14, 2016,
  Proceedings, Part II 14}.\hskip 1em plus 0.5em minus 0.4em\relax Springer,
  2016, pp. 102--118.

\bibitem{shafaei2016play}
A.~Shafaei, J.~J. Little, and M.~Schmidt, ``Play and learn: Using video games
  to train computer vision models,'' \emph{arXiv preprint arXiv:1608.01745},
  2016.

\bibitem{hanke2017generation}
T.~Hanke, A.~Schaermann, M.~Geiger, K.~Weiler, N.~Hirsenkorn, A.~Rauch, S.-A.
  Schneider, and E.~Biebl, ``Generation and validation of virtual point cloud
  data for automated driving systems,'' in \emph{2017 IEEE 20th International
  Conference on Intelligent Transportation Systems (ITSC)}.\hskip 1em plus
  0.5em minus 0.4em\relax IEEE, 2017, pp. 1--6.

\bibitem{fang2020augmented}
J.~Fang, D.~Zhou, F.~Yan, T.~Zhao, F.~Zhang, Y.~Ma, L.~Wang, and R.~Yang,
  ``{Augmented lidar simulator for autonomous driving},'' \emph{IEEE Robotics
  and Automation Letters}, vol.~5, no.~2, pp. 1931--1938, 2020.

\bibitem{fang2018simulating}
J.~Fang, F.~Yan, T.~Zhao, F.~Zhang, D.~Zhou, R.~Yang, Y.~Ma, and L.~Wang,
  ``{Simulating LIDAR point cloud for autonomous driving using real-world
  scenes and traffic flows},'' \emph{arXiv preprint arXiv:1811.07112}, vol.~1,
  2018.

\bibitem{manivasagam2020lidarsim}
S.~Manivasagam, S.~Wang, K.~Wong, W.~Zeng, M.~Sazanovich, S.~Tan, B.~Yang,
  W.-C. Ma, and R.~Urtasun, ``{Lidarsim: Realistic lidar simulation by
  leveraging the real world},'' in \emph{Proceedings of the IEEE/CVF Conference
  on Computer Vision and Pattern Recognition}, 2020, pp. 11\,167--11\,176.

\bibitem{lidarsim}
J.~Riordan, M.~Manduhu, J.~Black, A.~Dow, G.~Dooly, and S.~Matalonga, ``Lidar
  simulation for performance evaluation of uas detect and avoid,'' in
  \emph{2021 International Conference on Unmanned Aircraft Systems (ICUAS)},
  2021, pp. 1355--1363.

\bibitem{thrust}
\BIBentryALTinterwordspacing
{NVIDIA Cooperation}, ``{Thrust Prefix Sums} documentation,'' 2023. [Online].
  Available: \url{https://thrust.github.io/doc/group__prefixsums.html}
\BIBentrySTDinterwordspacing

\bibitem{cublas}
\BIBentryALTinterwordspacing
------, ``{cuBLAS} documentation,'' 2023. [Online]. Available:
  \url{https://docs.nvidia.com/cuda/cublas/}
\BIBentrySTDinterwordspacing

\bibitem{cuda}
\BIBentryALTinterwordspacing
------, ``{CUDA C/C++ Programming Guide},'' 2019. [Online]. Available:
  \url{https://docs.nvidia.com/cuda/cuda-c-programming-guide/}
\BIBentrySTDinterwordspacing

\bibitem{SORA}
\BIBentryALTinterwordspacing
{European Union Aviation Safety Agency}. {Specific Operations Risk Assessment
  (SORA)}. [Online]. Available:
  \url{https://www.easa.europa.eu/en/domains/drones-air-mobility/operating-drone/specific-category-civil-drones/specific-operations-risk-assessment-sora}
\BIBentrySTDinterwordspacing

\bibitem{DJI-L1-LiDAR}
\BIBentryALTinterwordspacing
DJI. {DJI L1 LiDAR}. [Online]. Available:
  \url{https://www.dji.com/uk/zenmuse-l1}
\BIBentrySTDinterwordspacing

\bibitem{Avia}
\BIBentryALTinterwordspacing
{Livox Technology Company}. {Livox Avia}. [Online]. Available:
  \url{https://www.livoxtech.com/avia}
\BIBentrySTDinterwordspacing

\bibitem{Optix}
\BIBentryALTinterwordspacing
{NVIDIA}. {NVIDIA OptiX™ Ray Tracing Engine}. [Online]. Available:
  \url{https://developer.nvidia.com/rtx/ray-tracing/optix}
\BIBentrySTDinterwordspacing

\bibitem{wang2019latte}
B.~Wang, V.~Wu, B.~Wu, and K.~Keutzer, ``Latte: accelerating lidar point cloud
  annotation via sensor fusion, one-click annotation, and tracking,'' in
  \emph{2019 IEEE Intelligent Transportation Systems Conference (ITSC)}.\hskip
  1em plus 0.5em minus 0.4em\relax IEEE, 2019, pp. 265--272.

\bibitem{sane}
H.~A. Arief, M.~Arief, G.~Zhang, Z.~Liu, M.~Bhat, U.~G. Indahl, H.~Tveite, and
  D.~Zhao, ``Sane: Smart annotation and evaluation tools for point cloud
  data,'' \emph{IEEE Access}, vol.~8, pp. 131\,848--131\,858, 2020.

\bibitem{aviaLatancy}
\BIBentryALTinterwordspacing
{LIVOX}. {Livox Introduces High Performance, Low Cost, Mass Market Lidar
  Sensors For L3/L4 Autonomous Driving Applications}. [Online]. Available:
  \url{https://www.livoxtech.com/news/4}
\BIBentrySTDinterwordspacing

\bibitem{pcdetPP}
\BIBentryALTinterwordspacing
{OpenPCDet}. [Online]. Available: \url{https://github.com/open-mmlab/OpenPCDet}
\BIBentrySTDinterwordspacing

\bibitem{openpcdet2020}
O.~D. Team, ``Openpcdet: An open-source toolbox for 3d object detection from
  point clouds,'' \url{https://github.com/open-mmlab/OpenPCDet}, 2020.

\bibitem{NMS}
J.~Hosang, R.~Benenson, and B.~Schiele, ``Learning non-maximum suppression,''
  in \emph{2017 IEEE Conference on Computer Vision and Pattern Recognition
  (CVPR)}, 2017, pp. 6469--6477.

\bibitem{losasso2004surface}
F.~Losasso, ``Surface reflection models,'' \emph{NVIDIA Corporation}, vol.~2,
  2004.

\end{thebibliography}
